\documentclass{article}

\usepackage{arxiv}

\usepackage[utf8]{inputenc} 
\usepackage[T1]{fontenc}    
\usepackage{hyperref}       
\usepackage{url}            
\usepackage{booktabs}       
\usepackage{amsfonts}       
\usepackage{nicefrac}       
\usepackage{microtype}      
\usepackage{graphicx}
\usepackage[numbers]{natbib}
\usepackage{doi}
\usepackage{amsmath}
\usepackage{siunitx}
\usepackage{graphicx}
\usepackage{array}
\usepackage{booktabs}
\usepackage{multirow}

\title{A pipeline for automated processing of Corona KH-4 (1962-1972) stereo imagery}

\date{} 					

\author{ {\hspace{1mm}Sajid~Ghuffar}\thanks{Corresponding author,          Email: sg313@st-andrews.ac.uk. Also affiliated with Department of Space Science, Institute of Space Technology, Islamabad, Pakistan} \\
	School of Geography and Sustainable Development \\
	University of St Andrews, St Andrews, UK  \\
	\And
	{\hspace{1mm}Tobias~Bolch} \\
	School of Geography and Sustainable Development \\
	University of St Andrews, St Andrews, UK\\
    \AND
    {Ewelina Rupnik} \\
	 LASTIG, Univ. Gustave Eiffel \\
	 ENSG, IGN, F-94160 Saint-Mande, France \\
	\And
    {Atanu Bhattacharya} \\
	Department of Remote Sensing and GIS\\
	JIS University, Kolkata, India \\
}

\begin{document}
\maketitle
\begin{abstract}
The Corona KH-4 reconnaissance satellite missions from 1962-1972 acquired panoramic stereo imagery with high spatial resolution of 1.8-7.5 m. The potential of 800,000+ declassified Corona images has not been leveraged due to the complexities arising from handling of panoramic imaging geometry, film distortions and limited availability of the metadata required for georeferencing of the Corona imagery. This paper presents Corona Stereo Pipeline (CoSP): A pipeline for processing of Corona KH-4 stereo panoramic imagery. CoSP utlizes a deep learning based feature matcher \textit{SuperGlue} to automatically match features point between Corona KH-4 images and recent satellite imagery  to generate Ground Control Points (GCPs). To model the imaging geometry and the scanning motion of the panoramic KH-4 cameras, a rigorous camera model consisting of modified collinearity equations with time dependent exterior orientation parameters is employed. The results show that using the entire frame of the Corona image, bundle adjustment using well-distributed GCPs results in an average standard deviation (SD) or $\sigma_0$ of less than $2$ pixels. We evaluate fiducial marks on the Corona films and show that pre-processing the Corona images to compensate for film bending improves the accuracy. We further assess a polynomial epipolar resampling method for rectification of Corona stereo images. The distortion pattern of image residuals of GCPs and y-parallax in epipolar resampled images suggest that film distortions due to long term storage as likely cause of systematic deviations. Compared to the SRTM DEM, the Corona DEM computed using CoSP achieved a Normalized Median Absolute Deviation (NMAD) of elevation differences of $\approx$ \SI{4}{\m} over an area of approx. \SI{4000}{\km \squared}. We show that the proposed pipeline can be applied to sequence of complex scenes involving high relief and glacierized terrain and that the resulting DEMs can be used to compute long term glacier elevation changes over large areas.
\end{abstract}


\section{Introduction}
The US Corona reconnaissance program consisted of a series of low-Earth orbit satellite missions. These missions were designed by the Central Intelligence Agency (CIA) for acquiring photographic images of primarily the former Soviet Union for strategic surveillance during the Cold War era \cite{macdonald1997corona}. Panoramic imagery from Corona missions is the first Earth observation dataset, which covers large part of the Earth's landmass.  The combination of high spatial resolution of up to \SI{1.8}{\m} along with stereoscopic coverage makes this data highly valuable in numerous applications \cite{galiatsatos2004assessment,altmaier2002digital}. In spite of that, the potential of this data has remained largely unexploited due to the difficulties in modeling of the panoramic geometry combined with the presence of film distortions and the limited availability of image metadata.

\begin{figure}[!t]
\centering
\includegraphics[width=0.6\linewidth]{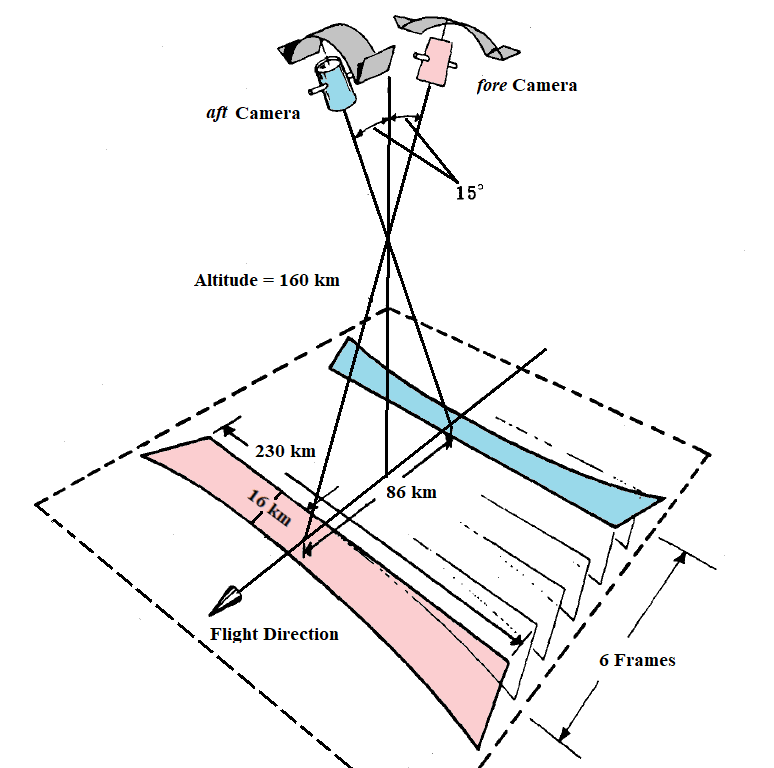}
\caption{Geometry of the Corona panoramic stereo imaging system (Reproduced and modified from \cite{NROJ3})}
\label{fig:CoronaGeomtery}
\end{figure}

The Corona spy satellite program was approved in the year 1958 and there were 102 Corona missions between 1960 and 1972. The data from these missions consisting of more than 800,000 images, were declassified in the year 1995 through a presidential order (Executive Order 12951). The code names for Corona series are KH-1, KH-2, KH-3, KH-4 KH-4A, KH-4B, where KH means \textit{Key-Hole}, a designation used for spy satellites \cite{Dashora07,macdonald1997corona}. In the KH-1 to KH-3 missions, the satellite imaging payload consisted of a single panoramic camera with ground resolution between 7.5-12 m. The KH-4 series consisted of a dual panoramic camera system with first mission in the year 1962. The dual camera system consisted of \textit{fore} (forward) and \textit{aft}  (backward) looking cameras with a $15^\circ$ off nadir view angle resulting in a  convergence of angle of 30$^\circ$ and baseline to height ratio of 0.54. These cameras consisted of a rotating lens and scan arm with a narrow slit that sequentially exposes a stationary film through a $70^\circ$ rotation of the scan arm (Figure \ref{fig:CoronaGeomtery}). The lens system of the \textit{fore} and \textit{aft} looking cameras rotated in opposite direction to minimize perturbations on the spacecraft.

The Corona panoramic images were captured on photographic film that was returned to the Earth using a re-entry capsule and recovered by the US Air Force.  The technical design details of the camera systems on board Corona mission are given in the declassified documents available from National Reconnaissance Office (NRO) \cite{NROIndex}. The Corona missions have undergone continuous design improvement since their inception and the documents and reports available from NRO provides valuable insights in to the design and working of these camera systems \cite{NROJ3,NROMural}. 

 The cameras in the Corona missions were designed by Itek Corporation and the main difference in the KH-4, KH-4A and KH-B cameras was the utilization of a rotating lens/scan arm in  KH-4B camera system instead of an oscillating scan arm in KH-4/4A camera systems \cite{madden1996corona}.  This provided more stability to the platform and allowed faster scanning rates, which enabled KH-4B satellites to be operated at a lower altitude and achieve a ground resolution of up to \SI{1.8}{\m}. There were 17 KH-4B missions with the first launch in 1967 and the final launch in 1972.  Table \ref{tab:CoronaTable} summarizes the main specifications of the KH-4 series missions. There were further variations in the image motion compensation system, camera filters, slit width and panoramic geometry reference data exposed on the film in the KH-4 series missions.

\begin{table}[]
\caption{Corona KH-4 mission and camera specifications \cite{NROJ3,NROMissionSt,JPL}. Due to a non circular orbit, the satellite altitude varies, which also changes the ground resolution and footprint.}  
\label{tab:CoronaTable}
\centering
\begin{tabular}{|c|c|c|c|  }
\hline 
 &  KH-4 & KH-4A  &  KH-4B    \\ 
\hline 
$\#$ of Missions & 26    &	 52    & 17 \\
Years  		   & 1962-1963 & 1963-1969	   & 1967-1972	      \\
Recovery Vehicles   & 1  &	2 &	2 \\
Orbit Perigee (km) & 200 & 180  & 150 \\
Camera Name         & Mural & J-1 & J-3 \\
Lens/Scan Arm Motion &  Reciprocating & Reciprocating  &  Rotating\\
Focal Length (mm)   & 609.6   &	609.6 &	609.6 \\
Stereo Angle (deg)  & 30  &	30 &	30 \\
Panoramic FOV (deg)  & 70  &	70 &	70 \\
Best Ground Res. (m) & 7.5 & 2.75  & 1.8 \\
Ground Footprint (km) & 20$\times$280 & 17$\times$232 & 14$\times$ 188\\
\hline
\end{tabular}
\end{table}

In this work we present CoSP: \textit{A pipeline for automated processing of Corona stereo imagery}, which implements a rigorous Corona camera model and automates the processing over the entire Corona image swath and in multi-image configurations. In addition to that, this paper contains following contributions: 
\begin{itemize}
\item We devise a scheme to automatically generate GCPs between Corona scenes and modern satellite imagery of different spatial resolutions, such as the images from Planet Labs nano-satellites and Landsat-7;
\item We experimentally show that the rigorous camera model with the time-dependent collinearity equations can achieve a SD of better than 2 pixels across the entire image footprint. We also show that estimated camera parameters are consistent with the Corona camera characteristics and the orbital parameters and different IMC mechanisms in KH-4A and KH-4B are also observed in the estimated parameters. 
\item We perform an investigation of the internal camera calibration through the reference data exposed to the film (e.g., rail holes, PG stripes). We propose a correction scheme and show that by removing the observed deviations, the accuracy in 3D can be improved;
\item We further show that the entire Corona stereo pair can be rectified with a residual y-parallax less one pixel but suspected film distortions lead to larger systematic y-parallax of up to 5 pixels in certain parts of the image;
\item We show a large-scale application of this pipeline by applying it to 24 Corona images, over a footprint of 160x200 km.  
\end{itemize}

\section{Related work}
\label{ch:relatedWork}
Due to the availability of images from the 1960s along with high spatial resolution, Corona images have enormous value in detecting long-term changes on the Earth's surface, such as changes of coastlines \cite{bayram2004coastline}, lakes   \cite{saruulzaya2016thermokarst} and urban areas \cite{noaje2012environmental}.
Most of the earlier work on Corona imagery, however,  has focused on applications in archaeology and cryosphere. Corona imagery has been used for identification and interpretation of archaeological features such as crop marks and built-up structures \cite{galiatsatos2004assessment,fowler2005detection,watanabe2017utilization}, computation of long term glacier area changes  \cite{bhambri2011glacier,schmidt2012changes,robson2018spatial,racoviteanu2015spatial}, estimation of glacier volume change and mass balance \cite{bolch2008planimetric,goerlich2017glacier,king2020six}, assessing area elevation and surface velocity changes of rock glaciers \cite{bolch2019occurrence, kaab2021inventory} and identification/mapping of glacial lakes \cite{bolch2011identification,raj2013remote}. \citet{bhattacharya2021high} have used DEMs generated from Corona KH-4 imagery for long term glacier mass balance estimation for several areas in High Mountain Asia. However, due to the manual work involved in their Corona image processing workflow, their work focused on certain smaller areas within the whole region. In this context, it should be mentioned that declassified images from KH-9 Hexagon (1971-1984) mapping cameras have been used for mapping of large areas and computation of glacier mass balance \cite{pieczonka2013heterogeneous,pieczonka2015region, dehecq2020automated,maurer2015tapping,maurer2019acceleration,king2019glacial}. In comparison to Corona, the processing of Hexagon KH-9 mapping camera imagery is less complicated because of frame camera geometry and availability of reseau grid for estimation of film distortions. The advantage of Corona imagery is higher spatial resolution in comparison to Hexagon (6-\SI{8}{\m}) as well as image acquisitions earlier in time \cite{maurer2015tapping}. This can extent the glacier mass balance time series, as well map the dynamics of slow flowing landforms such as rock glaciers at a high resolution over longer time period \cite{kaab2021inventory}.

The panoramic imaging geometry of the Corona camera systems differs from the frame camera geometry. As a result, different camera models and parameterization have been used in earlier studies to model the Corona panoramic geometry. \citet{sohn2004mathematical} evaluated three different camera models for KH-4B panoramic imagery. Their first model used modified collinearity equations, which included panoramic geometry transformation along with scan positional distortion and Image Motion Compensation (IMC) terms to model the difference between the frame camera geometry and the panoramic geometry. The scan positional distortion is caused by the movement of camera during the scan, while IMC compensates for the resulting image motion to avoid blurring. Their second model included the time dependent exterior orientation parameters in the collinearity equations to model the movement of the satellite over time as the rotating lens scans the scene. Their third camera model used the Rational Polynomial Coefficients (RPC) \cite{fraser2006sensor,tao2001comprehensive} to model the relationship between the image and the object coordinates. Their results showed that the first two models with modified collinearity equations resulted in a height RMSE of around \SI{4}{\m}  using GCPs covering an area of around \SI{560}{\km^2} (approx. 20\% area of entire stereo overlap), which was better than the height accuracy achieved using the RPC based model. \citet{shin2008rigorous} also used a camera model with time dependent exterior orientation parameters but only included the motion along the flight direction as an additional parameter. Their results showed a height RMSE of \SI{12.4}{m} on the check points spread over entire Corona stereo pair. The NASA Ames Stereo Pipeline (ASP) \cite{beyer2018ames} has implemented KH-4B panroamic camera model based on the model of \cite{shin2008rigorous} and motion compensation from \cite{sohn2004mathematical}.  \citet{lauer2019exploiting} has used a fisheye camera model for Corona imagery and obtained an average height accuracy of \SI{12.5}{\m} for the control points. \citet{jacobsen2020calibration} used the panoramic transformation terms in the perspective frame camera model and observed an average standard deviation of \SI{11.4}{pixels} using GCPs over multiple consecutive Corona stereo pairs. Multiple earlier studies \cite{bolch2008planimetric,goerlich2017glacier,bhattacharya2021high} have used RSG software (Remote Sensing Software Package Graz) developed by Joanneum Research Graz, for processing of Corona imagery to create DEMs for glacier volume change estimation. RSG Corona camera model is also based on panoramic coordinates transformation in the collinearity equations similar to the first model of \citet{sohn2004mathematical} but doesn't include IMC or scan positional distortion terms. These studies using RSG have reported on average 2.5 - 3 pixel image residuals using manually extracted GCPs.

Analyzing the earlier work on Corona image processing shows that further investigations are still required to assess the accuracy of the camera model over the entire panoramic image. Although \citet{sohn2004mathematical} presented a rigorous approach for modeling of Corona imagery, their evaluation was only performed over a relatively small area in comparison to the whole Corona image footprint.  Thus, additional evaluation is required to assess if similar accuracy can be achieved over the entire Corona image. There are also differences in the model parameterization in different studies. For example, RSG Corona camera model as well as in the work of  \citet{jacobsen2020calibration}, the camera model did not include scan motion distortion and the IMC terms. This is partially grounded by the fact that the IMC is designed to counteract the image motion. In a perfect scenario the effect of either is cancelled out. In reality, the image motion is compensated only to a certain degree, and the remaining residual causes systematic errors \cite{brown1965v}. 

In addition to the camera model, the incorporation of panoramic geometry reference data exposed on the film during the image acquisition has not been thoroughly evaluated in previous works. \cite{jacobsen2020calibration} has shown that the bending of the film has a significant magnitude and it should be corrected for achieving higher accuracy.

\section{Methodology}
\label{ch:Methodology}
\begin{figure}[!t]
\centering
  \includegraphics[width=0.8\linewidth]{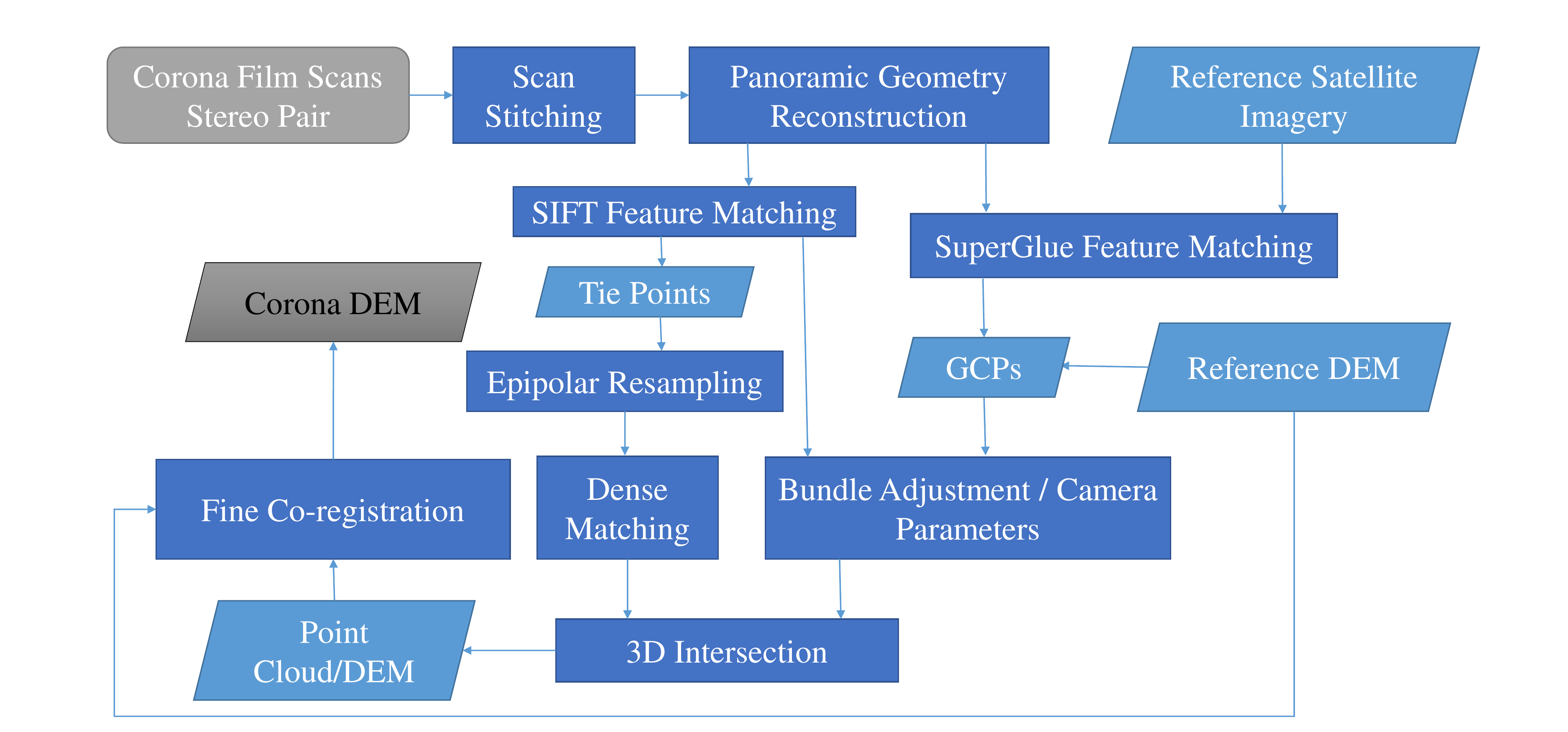}
  \caption{Workflow of the Corona Stereo Pipeline}
  \label{fig:workflow}
\end{figure}

The fore and aft looking camera of Corona KH-4 missions performs a $70^\circ$ scan in the across track direction with a rotating slit sequentially exposing a static film. The entire Corona image is not captured at one time instant as the platform is moving while the camera scans the scene in the across track direction. We therefore decided to use the modified collinearity equations with time dependent parameters to model the imaging geometry of Corona panoramic cameras based on the work of \citet{sohn2004mathematical}. This model has the advantage of modelling the motion of the platform during the scan with an explicit parameterization of the time dependent exterior orientation parameters. Main components of the Corona stereo pipeline are shown in the Figure \ref{fig:workflow}. The first steps involve recovery of the film geometry (including the individual scans' stitching and the reconstruction of the panoramic geometry) (Section~\ref{subsec:film_geometry}), followed by GCPs and tie points extraction (Section~\ref{subsec:gcp_extract}), the estimation of the camera model through bundle adjustment (Section~\ref{subsec:camera_model}), epipolar resampling (Section~\ref{subsec:epipolar_rectif}), dense matching, 3D intersection and fine coregistration (Section~\ref{subsec:dem_gen}). 

\subsection{Recovery of the film geometry}\label{subsec:film_geometry}

\paragraph{Film scans stitching}
The physical length of the film (across track direction) containing one panoramic scan is around \SI{745}{\mm}, while the film width (flight direction) is \SI{70}{\mm}. Due to its large size, the U.S. Geological Survey (USGS) scans the film in four parts with an overlap between the successive scans. The USGS allows film scanning at \SI{7}{\um} or \SI{14}{\um} resolution. We stitch the individual film scans together to generate the image of the whole film for further processing. The stitching operation is performed automatically by finding feature points in the overlapping regions of the scans and estimating a transformation (2D rotation and translation) to align them. The four scans are labeled by the USGS as \textit{a},\textit{b},\textit{c} and \textit{d}. We first stitch \textit{a} with \textit{b} and \textit{c} with \textit{d}, then the parts \textit{ab} and \textit{cd} are merged together.

\begin{figure}[!t]
\centering
  \includegraphics[width=\linewidth]{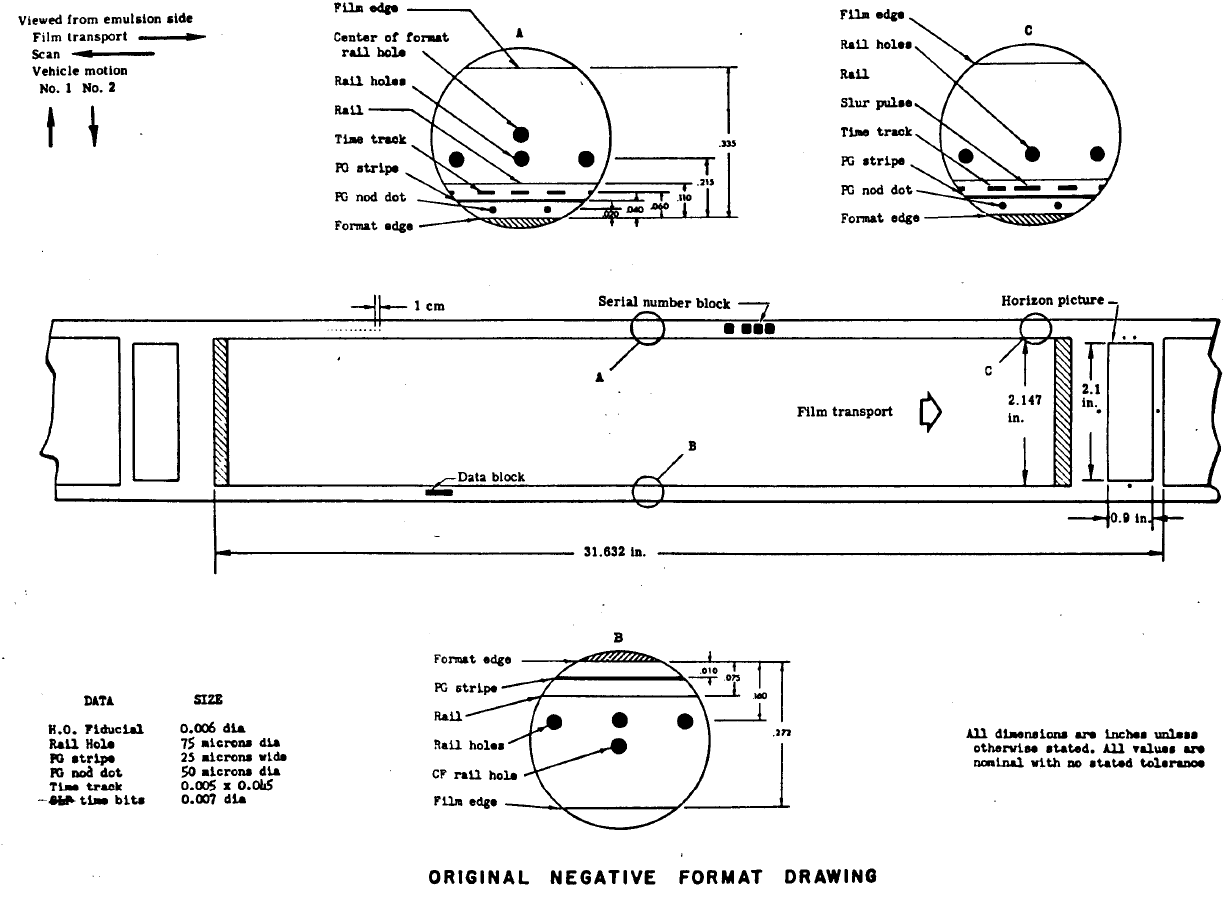}
  \caption{Schematic diagram of the film and the panoramic geometry reference data in KH-4B missions, (source: \cite{NROKH4B})}.
  \label{fig:FilmGeomtery}
\end{figure}

\paragraph{Fiducial marks}
During the image acquisition scan, additional data was exposed on the film to allow  reconstruction of the camera internal geometry and assessment of film distortions.  In the KH-4B missions the J-3 camera consisted of holes on the rails supporting the film, which produce a sequence of spots with a \SI{1}{\cm} spacing on both sides of the film length (Figure \ref{fig:FilmGeomtery}). The center of format included two rail holes on both sides of the film. In addition to the rail holes, there are Panoramic Geometry (PG) stripes running along the length of the film on both sides that can be used to reconstruct the panoramic geometry. 

\begin{figure}[!t]
  \includegraphics[width=\linewidth]{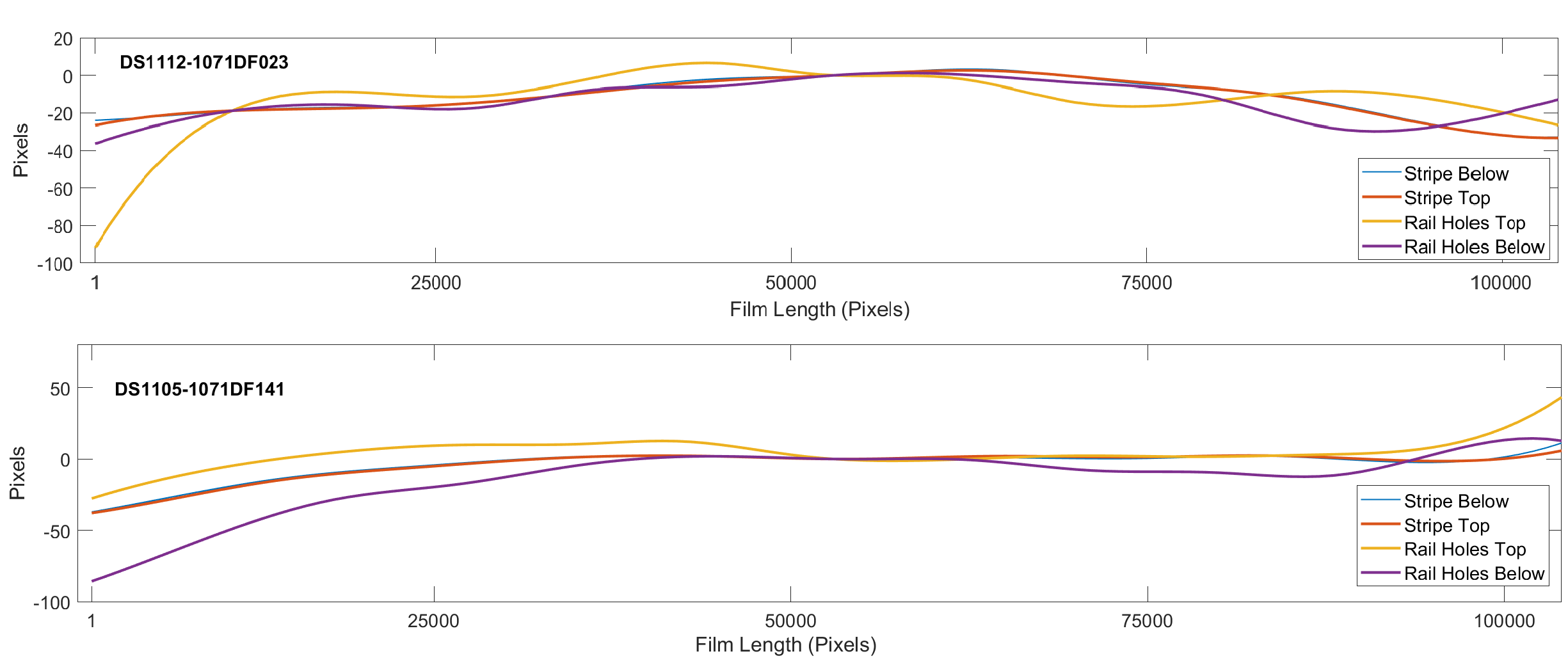}
  \caption{Position of the stripes and the rail holes above and below the image. These positions are plotted with reference to the respective locations corresponding to the center of the image. }
  \label{fig:Bend_DF141}
\end{figure}

We reconstructed the position of the rail holes and the PG stripes along the length of the film to assess the usability of this reference data for correcting the film bending and distortions. A certain number of positions of the PG stripes and the rail holes were extracted along the length of the film. These positions were then interpolated for each pixel along the length of the film. The position of the rail holes and the stripes in two Corona scanned images are shown in Figure \ref{fig:Bend_DF141}. We can observe that the positions of the rail holes along the two sides of the film unexpectedly deviate from each other and from the respective PG stripes. This may be due to the fact that the rail holes are more towards the ends of the film where distortions are higher in magnitude. This shows that the deviations in the rail holes are not necessarily representative of the film distortions towards the center of the image. On the other hand, the PG stripes which are comparatively closer to the imaged area reveal similar bending effects on either side of the film. Therefore, we use the location of the PG stripes to resample the image, imposing that the PG stripes become straight and parallel to each other. 

In CoSP, we extract and align the exposed image area from the entire scan as the scan also includes reference metadata and images from horizon cameras (Figure \ref{fig:FilmGeomtery}), which are not relevant for further processing. We first find an adaptive threshold to estimate exposed image pixels as the dark background has a low gray-value. Based on the estimated exposed image pixels, we find the centroid and principal components directions to rotate the image and align the length and width of the film scan along the x,y-axes of the image. The PG stripes are then extracted by finding the location of the edge pixels extracted using a canny edge detector at top and bottom of the image. The outliers are filtered out by median filtering the location of the edge pixels. The images are then resampled so that the PG stripes become straight and parallel to each other. Finally, we clip the \SI{1.5}{\cm} of the film on both ends along the length of the film due to focus variations. This focus anomaly is caused by the clearance required to avoid any collision between the rotating scan-head and the film rollers \cite{NRONSA}. The \textit{aft} camera images are rotated $180^\circ$ to align them with \textit{fore} camera images. Note that the rail holes and PG stripes are not available on the films of KH-4 and KH-4A missions and the above-mentioned correction can not be performed for images of these missions.

\subsection{GCP Generation}\label{subsec:gcp_extract}
In the earlier studies, the extraction of GCPs has been one of the most time consuming component of the Corona imagery processing pipeline. These GCPs were typically extracted by manually identifying features in Corona images and some reference satellite imagery and extracting the corresponding elevation values from a reference DEM. Automatic feature matching using SIFT \cite{Lowe:SIFT04} or other traditional features detectors is challenging because of  physical changes in the scene due to long temporal difference. The radiometric differences and variations in cast shadows in high relief terrain makes feature matching particularly difficult. Recent progresses in deep learning based feature matching has shown promising results in matching features with high perspective and radiometric difference. Such detectors have already been used for matching historical images \cite{maiwald2021automatic,zhang2021feature}.

We use SuperGlue \cite{sarlin2020superglue} a deep learning based architecture for feature extraction and matching to compute point matches between Corona imagery and reference satellite imagery. The SuperGlue architecture consists of a Convolutional Neural Network (CNN) based interest point detector and descriptor called SuperPoint \cite{detone2018superpoint}, which uses VGG \cite{simonyan2014very} style encoder. It uses a graph neural network to perform matching of the features extracted by the CNN and implicitly learns their spatial relationship within one image and across a pair with the help of the attention mechanism. The attention mechanism serves as a spatial filter and is learned end-to-end. The SuperGlue like any other CNN based architecture is typically trained on relatively low resolution images. The default image resolution for SuperGlue is  $640 \times 480$ and we found in our experiments that it provides reasonable results up to three times the default resolution i.e, $1920 \times 1440$. The Corona image has a much larger size i.e. $\sim 106000 \times 8000$. So, we divide it in to patches of size $1920 \times 1440$ and extract similar geographic location patch from references satellite imagery to match the feature points. As the resolution of KH-4B images is around 2-3 \SI{}{\m}, we use reference satellite imagery at similar spatial resolution to match feature points. We used a mosaic of orthorectified PlanetScope images \cite{PlanetLabs} with \SI{3}{\m} spatial resolution for feature matching with the Corona imagery.

As high resolution reference satellite imagery may not be available for larger areas (here, we refer to 1-5\SI{}{\m} spatial resolution as high resolution imagery), we also experimented with feature matching on freely available medium resolution satellite imagery like Landsat-7 ETM+. When using medium resolution imagery, we resample the Corona imagery to have approx. similar spatial resolution as the ETM+ panchromatic band, i.e.,  \SI{15}{\m}. Specifically, we divide Corona image into tiles of $ \sim 10600 \times 8000$ and resize it to $1920 \times 1440$ for input to SuperGlue. 

We use the geographic locations of the image corners specified in the Corona image metadata to extract overlapping tiles from the Corona and reference satellite imagery. The locations of the image corners are derived from the satellite orbit, and can deviate several kilometers from their real positions in a typical scenario or up to \SI{100}{\km} in certain cases. If this deviation is larger than the size of the tile,  feature matching will not work as the tiles from Corona and reference satellite imagery will have no overlap. Therefore, we adopt a coarse-to-fine strategy. We first find the overlapping zones with low resolution images over an extended area. Then we use the refined estimates of the Corona footprint to recover precise correspondences in a localised area using higher resolution images.

\subsection{Camera Model}\label{subsec:camera_model}
\begin{figure}[]
  \includegraphics[width=\linewidth]{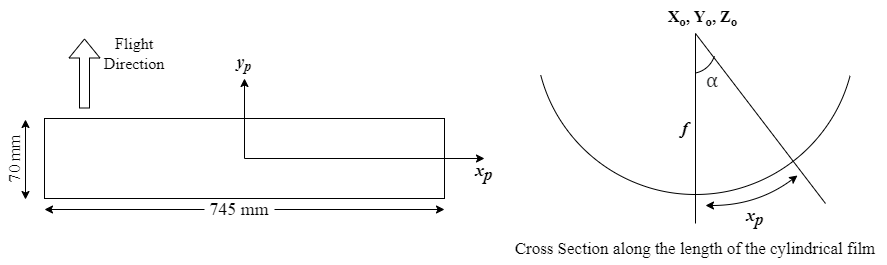}
  \caption{The panoramic image coordinate system and the angle $\alpha$. }
  \label{fig:FilmCoord}
\end{figure}

The scene is exposed to the film through a slit, which rotates along with the lens. The scan angle $\alpha$ is the angle of the slit and is given by

\begin{equation}
\alpha = \frac{x_p}{f} 
\end{equation}

where $f$ is the focal length and $x_p$ is the panoramic photo coordinate along the length of film  (Figure \ref{fig:FilmCoord}). The exterior orientation parameters are modeled as first order time dependent parameters

\begin{equation}
X_{0t} = X_0 + X_{01} \cdot t ,  
\end{equation}

\begin{equation}
Y_{0t} = Y_0 + Y_{01} \cdot t ,  
\end{equation}

\begin{equation}
Z_{0t} = Z_0 + Z_{01} \cdot t , 
\end{equation}

\begin{equation}
\omega_{0t} = \omega_0 + \omega_{01} \cdot t , 
\end{equation}

\begin{equation}
\phi_{0t} = \phi_0 + \phi_{01} \cdot t ,  
\end{equation}

\begin{equation}
\kappa_{0t} = \kappa_0 + \kappa_{01} \cdot t . 
\end{equation}

Here, $X_{01},Y_{01},Z_{01}$ represent the motion of the camera  and $\omega_{01}, \phi_{01},\kappa_{01}$ represent the angular motion along the three axis during the panoramic scan with $t\in [0,1]$.  Considering that the width of the slit to be small, the geometry of the image formed at any position of the slit is given by,
\begin{equation}
\begin{bmatrix} 0 \\ y_p+ y_{IMC} \\ -f \end{bmatrix} = s R_{\alpha} R \begin{bmatrix} X-X_{0t} \\ Y-Y_{0t} \\ Z-Z_{0t}\end{bmatrix}  ,
\label{eq:Col}
\end{equation}

where $R_{\alpha}$ is, 
\begin{equation}
R_{\alpha}=\begin{bmatrix}  cos \alpha & 0 & sin \alpha   \\ 0 & 1 & 0 \\    -sin \alpha & 0  &cos \alpha   \end{bmatrix} ,
\end{equation}

and the IMC term is given as, 
\begin{equation}
 y_{IMC} =- \frac{V f}{H \delta} sin(\alpha) cos(\omega_0),
\end{equation}
where, $V$ is the satellite velocity, $H$ is altitude of the satellite and $\delta$ is the scan angular velocity. The image motion caused by a moving platform is directly related to V/H ratio \cite{kawachi1963image,Itek2,brown1965v}. Therefore, as the orbital altitude varies the observed image motion also varies. In J-3 camera, the image motion was compensated by rotating the lens in the direction opposite to the image motion, while in the earlier J-1 system the lens was moved along the image y-axis to compensate image motion \cite{NROJ3}. The image motion compensation produces an S-shape distortion across the image length \cite{Itek2}.

{By rearranging the terms of Equation~\eqref{eq:Col} we obtain,}
\begin{equation}
R_{\alpha}^T \begin{bmatrix} 0 \\ y_p+ y_{IMC} \\ -f \end{bmatrix} = s  R \begin{bmatrix} X-X_{0t} \\ Y-Y_{0t} \\ Z-Z_{0t}\end{bmatrix}  ,
\label{eq:CamMod}
\end{equation}
{and finally,}
\begin{equation}
\begin{bmatrix} f sin \alpha \\ y_p+ y_{IMC} \\ -f cos \alpha \end{bmatrix} = s R \begin{bmatrix} X-X_{0t} \\ Y-Y_{0t} \\ Z-Z_{0t} \end{bmatrix}  .
\end{equation}

Let,
\begin{equation}
 N_x =  r_{11}(X-X_{0t}) + r_{12}(Y-Y_{0t}) + r_{13}(Z-Z_{0t}) 
\end{equation}
\begin{equation}
 N_y = {r_{21}(X-X_{0t}) + r_{22}(Y-Y_{0t}) + r_{23}(Z-Z_{0t}) } 
\end{equation}
\begin{equation}
 N_z =  r_{31}(X-X_{0t}) + r_{32}(Y-Y_{0t}) + r_{33}(Z-Z_{0t}) 
\end{equation}
where $r_{ii}$ are the elements of the rotation matrix $R$. Division of $1^{st}$ and the $2^{nd}$ row with the last row cancels out the scale factor and gives the following equations, 
\begin{equation}
 tan \alpha =  - \frac{N_x}{N_z},
\end{equation}

\begin{equation}
y + y_{IMC} = -f cos \alpha  \frac{N_y}{N_z},
\end{equation}

The final image panoramic coordinates $(x_p,y_p)$ being, 

\begin{equation}
x_p  = f tan^{-1}  ( -\frac{N_x}{N_z})
\end{equation}

\begin{equation}
y_p  = - y_{IMC} - f cos\alpha  \frac{N_y}{N_z} 
\end{equation}

There are 13 unknown camera parameters in the model: 6 parameters for camera position $X_0,Y_0,Z_0,X_{01},Y_{01},Z_{01}$, six parameters for camera orientation $\omega_0,\phi_0,\kappa_0,\omega_{01},\phi_{01},\kappa_{01}$ and one parameter for the IMC i.e. $\frac{V}{H \delta}$.  We keep the focal length fixed in the adjustment due to the correlation with the camera altitude parameter \cite{zhou2020simulation}. The approximate values for $\omega_0$ were set to  $\{-15^\circ, 15^\circ\}$ for the fore and aft looking cameras respectively, $\kappa_0$ as $0^\circ$, camera altitude $Z_0$ is set to \SI{170}{\km} and  the time dependent parameters are initialized as zero. These parameters are then optimized in bundle  adjustment using GCPs as well as tie points between the stereo pairs. The bundle adjustment is performed in an Earth Centered Cartesian frame.

\subsection{Stereo Rectification}\label{subsec:epipolar_rectif}
\begin{figure}[!t]
  \includegraphics[width=\linewidth]{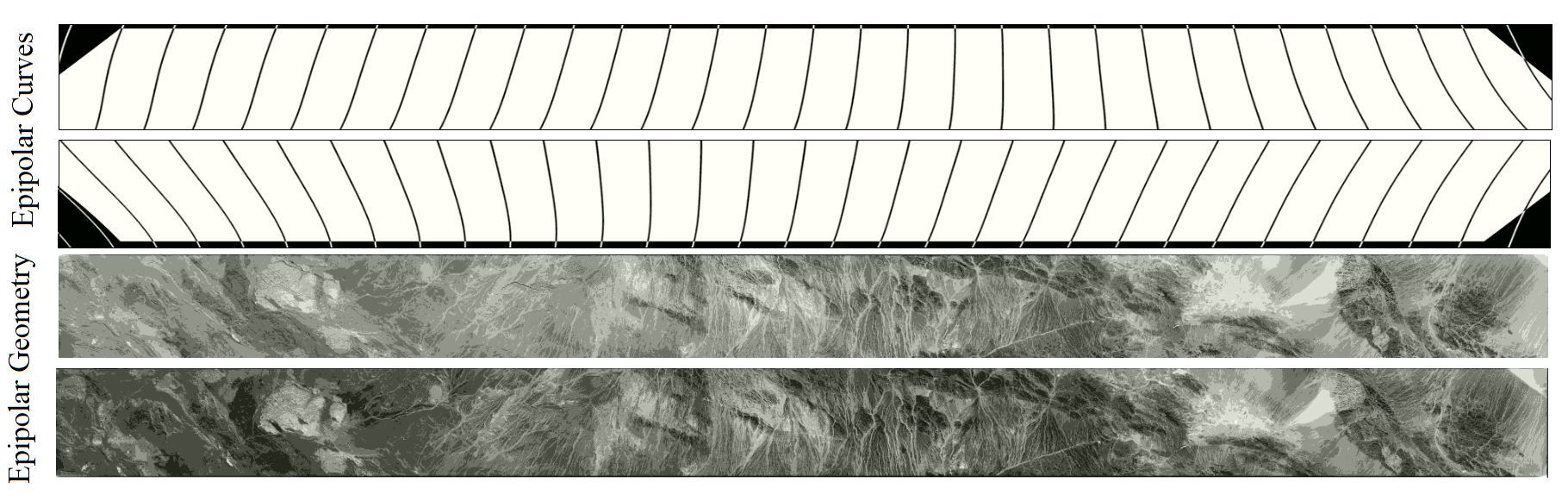}
  \caption{Epipolar curves in {the native geometry of} a Corona stereo pair {(top) and the resulting epipolar rectification (bottom)}. These curves are exaggerated for visualization. {Note that the curves are globally vertical and they will be rotated by $\approx 90^{\circ}$ prior to estimating the rectifying polynomials.} }
  \label{fig:EpipolarLines}
\end{figure}

We follow the stereo rectification algorithm given in \cite{deseilligny:hal-02968078} and implemented in MicMac~\cite{pierrot2014micmac}. This epipolar rectification is generic in the sense that it makes no assumption on the image pair camera model, as long as the image-to-ground projection functions are \textit{smooth}. {It is therefore suitable to uncommon image geometries such as that of the Corona images}. The algorithm consists of four stages: (1) estimation of the global epipolar directions in either image of the stereo-pair (see Figure~\ref{fig:EpipolarLines} (top)); (2) generation of a set of virtual correspondences between images using the known projection functions; (3) estimation of the rectifying polynomials in a rotated image coordinate frame where the epipolar curves are approximately horizontal; (4) computation of the final rectifying functions as a composition of the rotations computed in step 1 and the polynomials computed in step 3. If the camera geometry and the projection functions are unknown (e.g., due to lack of metadata or when handling challenging camera geometries such as Corona), the image correspondences can be replaced with image features extracted with image processing algorithms such as SIFT \cite{Lowe:SIFT04} or SuperGlue \cite{sarlin2020superglue}. Note that to obtain unambiguous rectifications using this variant, the 3D scene should not be flat. For more details, readers are referred to the original publication \cite{deseilligny:hal-02968078}. Here, we use a 4th degree polynomial to approximate the epipolar curves.

\subsection{DEM Generation and DEM Coregistration}\label{subsec:dem_gen}

We compute the dense correspondences between the stereo rectified image pairs with an implementation of the Semi Global Matching (SGM) \cite{hirschmuller2007stereo}. SGM approximates the 2D disparity cost as 1D cost along several paths, and has been widely used for DSM generation from aerial and satellite imagery.

Once the dense point correspondences and the camera parameters are known, 3D triangulation of the corresponding points is performed. Rearranging Eq. \ref{eq:CamMod}.  gives the following two equations for each image observation:
\begin{equation}
\begin{bmatrix} R_1 + tan(\alpha)R_3 \\ R_2 + \frac{y+y_{IMC}}{fcos(\alpha)} R_3 \end{bmatrix}  
\begin{bmatrix} X \\ Y \\ Z \end{bmatrix}  =
\begin{bmatrix} R_1 + tan(\alpha)R_3 \\ R_2 + \frac{y+y_{IMC}}{fcos(\alpha)} R_3)\end{bmatrix} 
\begin{bmatrix} X_{0t} \\ Y_{0t} \\ Z_{0t} \end{bmatrix}.
  \label{eq:ForwIntersect}
\end{equation}

Here, $R_1, R_2$ and $R_3$ are the 1$^{st}$, 2$^{nd}$ and 3$^{rd}$ rows of the rotation matrix respectively.
Given the four equations, a least squares minimization is performed to estimate the 3D coordinates for each point. This leads to a dense 3D point cloud, which is then interpolated into a raster DEM at the required spatial resolution. 

In practice, the GCPs computed automatically may not be well distributed across the image. This leads to a systematic deviation between the Corona DEM and the reference DEM.  Even with well distributed GCPs, film distortions can lead to systematic deviations between the two DEMs (see Figure \ref{fig:DEMDif}). In applications requiring DEM differencing, a fine coregistration between the two DEMs is necessary to produce unbiased results. Therefore, to compensate for these systematic deviations, we perform a fine coregistration of Corona DEM and the reference DEM in a post-processing step. Concretely, we divide the Corona DEM into smaller tiles and estimate a 3D affine transformation using  Least Square Matching (LSM) \cite{ressl2011applying}. We use a tile size of approx. \SI{400}{\km \squared} and estimate the parameters of the transformation individually for each tile. To ensure smooth transition between neighbouring tiles we keep a certain amount of overlap between them. 

We have implemented the camera model and bundle adjustment in MATLAB. Implementation of dense matching, tie point extraction and epipolar resampling is based on MicMac, while GCP extraction uses SuperGlue's original inference and evaluation script, and the DEM coregistration uses LSM module of OPALS \cite{pfeifer2014opals}. All these modules are executed from a script written in MATLAB. 

\section{Data and Study Areas}
\begin{figure}[!t]
  \includegraphics[width=\linewidth]{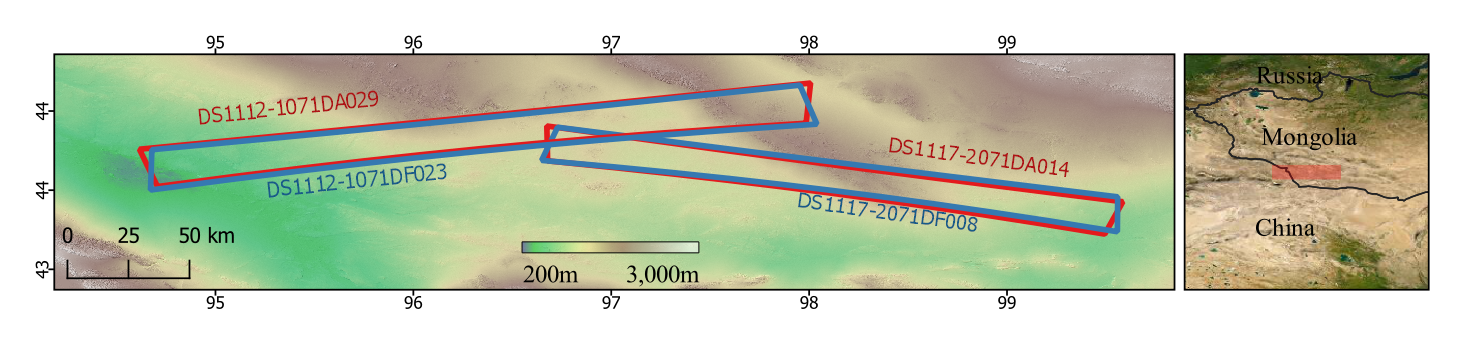}
   \includegraphics[width=\linewidth]{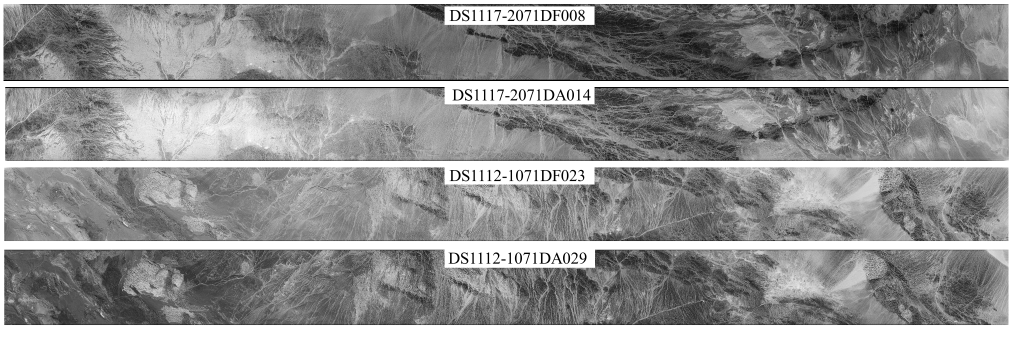}
  \caption{Corona scenes and corresponding footprints of the stereo pairs DS1117-2071DF008 DS1117-2071DA014 and  DS1112-1071DF023 DS1112-1071DA029 over an area bordering Mongolia and China }
  \label{fig:Mongolia_AOI}
\end{figure}

We applied and evaluated CoSP using three different case studies (see Table~\ref{tab:DataTable}). We first focus on the evaluation of camera model and select a study site for which well distributed GCPs can be automatically extracted. We searched for an area with distinctive texture and minimal natural or anthropogenic changes over the past decades. Furthermore, we restricted our search to areas with shallower slopes as high relief terrain is associated with larger elevation errors in the global DEMs, which we use in our study to extract the GCPs' elevation values.  Our visual search revealed that area bordering Mongolia and China is a suitable for this purpose. Therefore, we chose two Corona stereo pairs over an area bordering China and Mongolia for the evaluation of the camera model and the corresponding Corona DEM (Figure \ref{fig:Mongolia_AOI}). The two image pairs belong to mission no. 1112 and 1117 and the image acquisition dates are 23-11-1970 and 30-05-1972, respectively. The images were downloaded from the USGS Earth Explorer. The elevation range in the subject area is between 300-2500 \SI{}{\m}. We use SRTM 1 Arc-Second Global DEM v3 (DOI: 10.5066/F7PR7TFT) for extracting the elevation values and comparison with the Corona DEM. A mosaic of orthorectified PlanetScope imagery of this area is used for extraction of GCPs using SuperGlue. It is pertinent to mention that the GCPs also have limited accuracy as they are derived from PlanetScope imagery (for planimetric coordinates) and \SI{30}{\m} resolution reference DEM (for elevation coordinates).  

Our second focus is the evaluation of the proposed pipeline in estimating glacier change. To do that, we chose an area around Mount Everest in the Himalaya because of the availability of high resolution Cartosat DEM (\SI{10}{\m}) and Corona DEM computed using RSG. The latter allows us to compare the DEMs computed from our rigorous model, and the approximated variant of the Corona model implemented in RSG. The glaciers in the Himalaya are of high interest due to accelerating mass loss \cite{king2020six}. We use a Corona stereo pair from mission no. 1108 dated 18-12-1969 (Table \ref{tab:DataTable}) to create Corona DEM. Here, we use \SI{30}{\m} ALOS PRISM DEM (AW3D) \cite{tadono2014precise} as a reference DEM because of the bias in SRTM DEM due to snow penetration of radar signal. 

Finally, to evaluate the potential of large scale mapping using CoSP, we use 12 successive stereo pairs over Central Pamir in Tajikistan from mission 1104, dated 18-08-1968 (Table \ref{tab:DataTable}), covering a footprint of approx. $160\times200$ km. This sequence of images are suitable for large scale glacier change estimation due to late summer acquisition and cloud free coverage.  

\begin{table*}[htbp]
  \centering
  \caption{Corona scenes used for each study site and the corresponding reference DEMs and satellite imagery used for GCPs}
    \begin{tabular}{|c|c|c|c|c|}
    \toprule
    \multicolumn{1}{|c|}{\textbf{Study}} & \textbf{Corona Scenes} & \multicolumn{1}{c|}{\textbf{Date}} & \multicolumn{1}{|c|}{\textbf{Imagery}} & \multicolumn{1}{c|}{\textbf{Reference}} \\
    \multicolumn{1}{|c|}{\textbf{Site}} & \multicolumn{1}{c|}{} &        & \multicolumn{1}{c}{\textbf{for GCPs}} & \multicolumn{1}{|c|}{\textbf{DEM}} \\
    \midrule
    \multicolumn{1}{|c|}{\multirow{4}[2]{*}{Mongolia}} & DS1117-2071DF008 & \multicolumn{1}{c|}{\multirow{2}[1]{*}{30-05-1972}} & \multicolumn{1}{c|}{\multirow{4}[2]{*}{PlanetScope }} & \multicolumn{1}{c|}{\multirow{4}[2]{*}{SRTM }} \\
           & DS1117-2071DA014 &        &        &  \\
           & \multicolumn{1}{c|}{DS1112-1071DF023} & \multicolumn{1}{c|}{\multirow{2}[1]{*}{23-11-1970}} &        &  \\
           & DS1112-1071DA029 &        &        &  \\
    \midrule
    \multicolumn{1}{|c|}{\multirow{2}[2]{*}{Everest}} & DS1108-2217DA070       & \multicolumn{1}{c|}{\multirow{2}[2]{*}{18-12-1969}} & \multicolumn{1}{c|}{\multirow{2}[2]{*}{Landsat-7}} & \multicolumn{1}{c|}{ALOS } \\
           & DS1108-2217DF064 &        &        & \multicolumn{1}{c|}{Cartosat} \\
    \midrule
    \multicolumn{1}{|c|}{\multirow{4}[2]{*}{Pamir}} & DS1104-2169DF092- & \multicolumn{1}{c|}{\multirow{4}[2]{*}{18-08-1968}} & \multicolumn{1}{c|}{\multirow{4}[2]{*}{Landsat-7 }} & \multicolumn{1}{c|}{\multirow{4}[2]{*}{ALOS}} \\
           & DS1104-2169DF103 &        &        &  \\
           & DS1104-2169DA098- &        &        &  \\
           & DS1104-2169DA109 &        &        &  \\
    \bottomrule
    \end{tabular}%
  \label{tab:DataTable}%
\end{table*}%


\section{Results}
\label{ch:Results}
\subsection{Accuracy of the Camera Model (\textit{Mongolia Data})}
\begin{figure}[]
  \includegraphics[width=\linewidth]{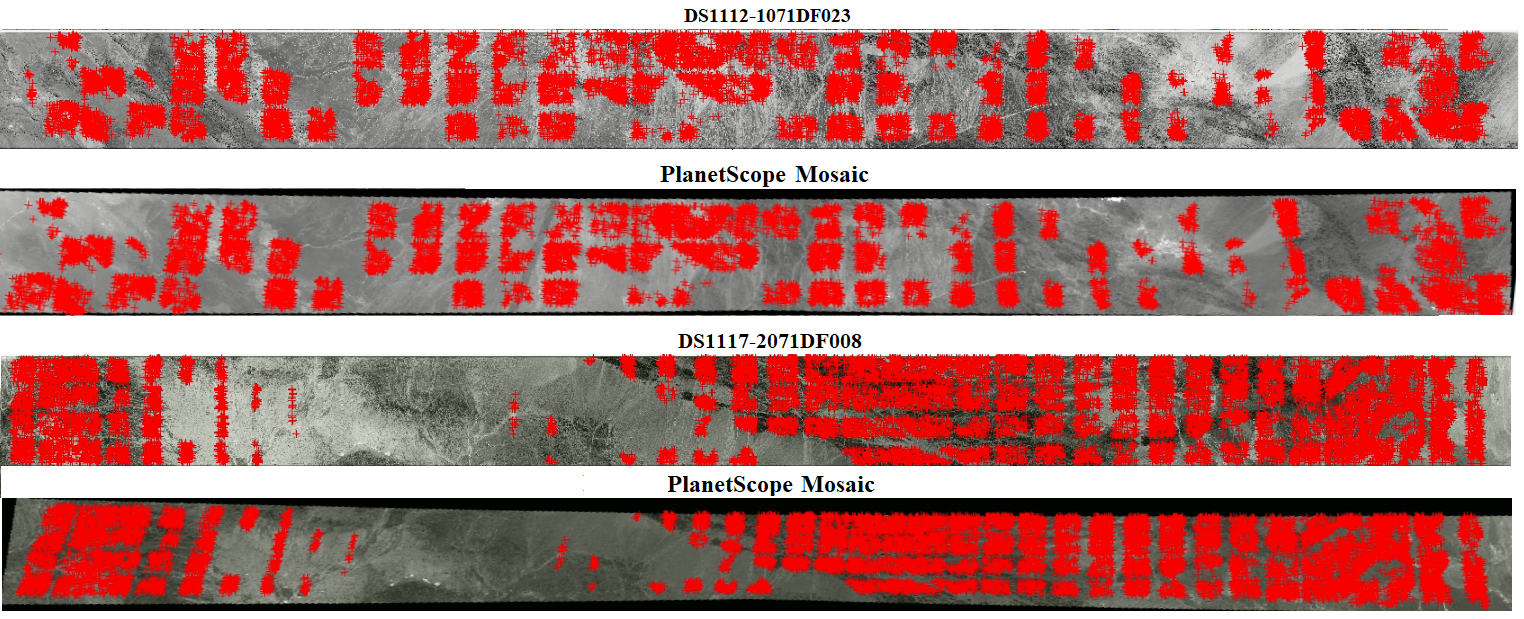}
  \caption{Point matches between a Corona images and multi satellite PlanetScope imagery mosaic computed by SuperGlue. The grid type pattern of the feature points is due to tile based matching.}
  \label{fig:FeatMong}
\end{figure}

\paragraph{{Accuracy on Check Points}} SuperGlue extracted large number of features points between the Corona images and the high resolution satellite imagery (see Figure \ref{fig:FeatMong}). Fewer number of points were detected in areas without texture as well as the areas with surface changes. These feature points along with the corresponding height values derived from the SRTM DEM form the GCPs and half of them are used as check points to evaluate the accuracy of the camera model. The average standard deviation ($\sigma_0$) of bundle adjustment is $\sim 2$ pixels for both the image pairs (Table \ref{tab:resultsMong}). The $\sigma_0$ shows an improvement of $10-20\%$ when bending correction is applied. The RMSE error of the checkpoints for the two image pairs are $[4.85,3.81,5.79] $ and  $[6.63,4.23,8.42]$ (m)  (Table \ref{tab:resultsMong}). As with the $\sigma_0$, the RMSE decreases when film bending correction is performed. 

\begin{table*}[htbp]
  \centering
  \caption{Accuracy of the bundle adjustment and RMSE of the check points for the two image pairs shown in Figure \ref{fig:Mongolia_AOI}. The RMSE is computed in the corresponding UTM zone.}
    \begin{tabular}{|ccccc|}
    \toprule
    \multirow{2}[2]{*}{\textbf{Image Pair  }} & \textbf{$\sigma_0$} & \textbf{RMSE X   } & \textbf{RMSE Y } & \textbf{RMSE Z } \\
          & \textbf{(pixels)} & \textbf{(m)} & \textbf{(m)} & \textbf{(m)} \\
    \midrule
    \textbf{Bending Correction} &       &       &       &  \\
    DS1112-1071DF023 & \multirow{2}[0]{*}{1.71} & \multirow{2}[0]{*}{4.85} & \multirow{2}[0]{*}{3.81} & \multirow{2}[0]{*}{5.79} \\
    DS1112-1071DA029 &       &       &       &  \\
    DS1117-2071DF008 & \multirow{2}[1]{*}{1.90} & \multirow{2}[1]{*}{6.63} & \multirow{2}[1]{*}{4.23} & \multirow{2}[1]{*}{8.42} \\
    DS1117-2071DA014 &       &       &       &  \\
    \midrule
    \textbf{Without Correction} &       &       &       &  \\
    DS1112-1071DF023 & \multirow{2}[0]{*}{1.89} & \multirow{2}[0]{*}{5.84} & \multirow{2}[0]{*}{4.09} & \multirow{2}[0]{*}{9.92} \\
    DS1112-1071DA029 &       &       &       &  \\
    DS1117-2071DF008 & \multirow{2}[1]{*}{2.43} & \multirow{2}[1]{*}{7.39} & \multirow{2}[1]{*}{4.75} & \multirow{2}[1]{*}{15.11} \\
    DS1117-2071DA014 &       &       &       &  \\
    \bottomrule
    \end{tabular}%
  \label{tab:resultsMong}%
\end{table*}%

\begin{figure}[]
  \includegraphics[width=\linewidth]{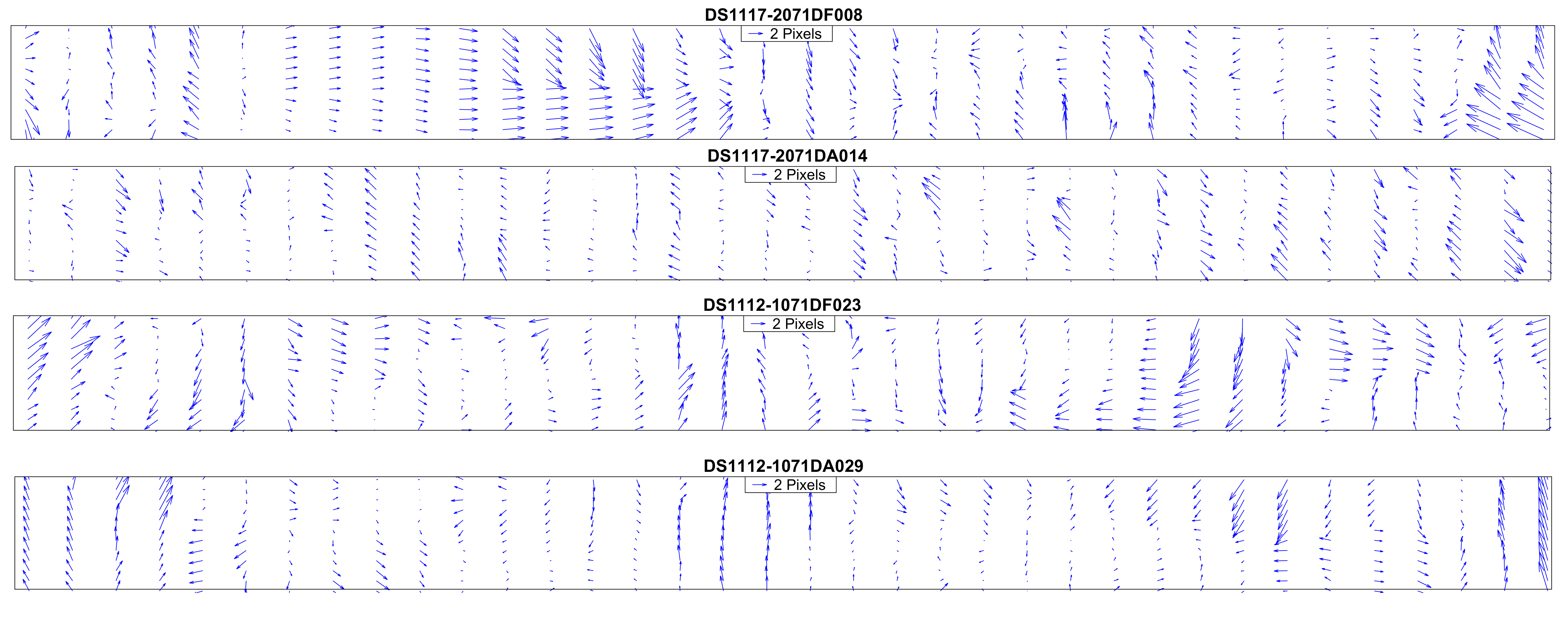}
  \caption{Image residuals of the GCPs after resection are interpolated over a regular grid. These images have been corrected for film bending. }
  \label{fig:DistF023}
\end{figure}

\paragraph{{Residual systematic errors}} In order to assess the residual distortions in the image,  residuals are computed from back projection of the GCPs and interpolated over a regular grid (Figure \ref{fig:DistF023}). Image residuals show higher magnitude in some parts of the image especially towards the edges of the film. These image residuals can mainly be attributed to film distortions that occurred during the mission and long term storage of the film.

\begin{table*}[htbp]
  \centering
  \caption{The estimated camera parameters are given according to the geographic coordinate system with the position of the camera given as longitude, latitude and elevation.}
    \begin{tabular}{|c|c|c|c|c|}
    \toprule
    \multirow{2}[2]{*}{\textbf{Parameters}} & \textbf{DS1112-} & \textbf{DS1112-} & \textbf{DS1117-} & \textbf{DS1117-} \\
          & \textbf{1071DF023} & \textbf{1071DA029} & \textbf{2071DF008} & \textbf{2071DA014} \\
    \midrule
    $X_0$  (Lon.) & 96.24 & 96.39 & 98.26 & 98.04 \\
    $Y_0$  (Lat.) & 44.59 & 43.70 & 44.31 & 43.54 \\
    $Z_0$  (km) & 187.27 & 186.77 & 162.21 & 162.89 \\
    $ X_{01}$ (km) & 0.06  & 1.20  & -1.53 & 0.41 \\
    $Y_{01}$ (km) & -2.75 & -2.86 & -2.59 & -3.10 \\
    $Z_{01}$ (km) & -0.41 & -0.03 & 0.15  & -0.14 \\
    $\omega_0$ (deg) & -15.20 & 15.72 & -15.52 & 14.68 \\
    $\phi_0$ (deg) & -1.56 & 1.46  & 3.71  & -2.25 \\
    $\kappa_0$ (deg) & 5.69  & 5.74  & -10.19 & -10.64 \\
    $\omega_{01}$ (deg) & 0.83  & 0.94  & 0.76  & 1.09 \\
    $\phi_{01}$  (deg) & -0.04 & -0.03 & -0.51 & -0.23 \\
     $\kappa_{10}$ (deg) & 0.00  & -0.23 & 0.33  & 0.02 \\
    $\frac{V}{H\delta}$ & 0.0025 & -0.0001 & -0.0002 & 0.0002 \\
    \bottomrule
    \end{tabular}%
  \label{tab:CameraParam}%
\end{table*}%

\paragraph{{Camera parameters}} Overall, the estimated camera parameters correspond relatively well to the expected values (see Table \ref{tab:CameraParam}). The estimated position and the orientation of the satellite is consistent with the orbital parameters of the respective acquisitions and stereo geometry. The main component of the estimated satellite motion during the scan is along north direction i.e. $Y_{01}$, whose estimated value ranges from 2.59-3.10 \SI{}{\km}. The expected value of motion along flight direction for a  scanning time of 0.36 sec is \SI{2.8}{\km} (velocity = \SI{7.7}{\km \per \s}). The expected values of other time dependent orientation parameters will be close to zero if we assume a stable satellite attitude during the acquisition. 

The specifications of the J-3 camera states that the IMC in KH-4B missions was implemented by rotating the camera in a direction opposite to the flight direction, while in the earlier Corona missions IMC was implemented by translating the lens relative to the film \cite{NROJ3}. This may explain why:  1) The estimated value of the IMC term $\frac{V}{H\delta}$ is significantly smaller than the expected value of 0.014 (V= \SI{7.7}{\km \per \s}, H = \SI{170}{\km}, $\delta$= \SI{3.3}{\radian \per \s}) as both image pairs are from KH-4B missions and 2) A value between 0.76-1.09 deg for the parameter $\omega_{01}$ because the rotation to compensate camera motion will be observed in $\omega_{01}$. Given a scan time of 0.36 sec, the \SI{7.7}{\km \per \s} velocity of the satellite will lead to a change of approx. 0.9 deg angle for an object point at the center of the format. Thus, the estimated $\omega_{01}$ is consistent with the rotation required to compensate the image motion. To further assess the effect of IMC mechanism on the estimated parameters, we estimated camera parameters for different KH-4A scenes using GCPs generated from {the scheme introduced in Section~\ref{subsec:gcp_extract}}. In majority of the images, the estimated IMC term has a value closer to the value of 0.014 (Table \ref{tab:IMCKH4}).

\begin{table*}[htbp]
  \centering
  \caption{Estimated time dependent orientation parameters and IMC for KH-4A scenes}
    \begin{tabular}{|c|c|c|c|c|c|c|c|}
    \toprule
    \textbf{Image} & \textbf{$X_{01}$} & \textbf{$Y_{01}$} & \textbf{$Z_{01}$} & \textbf{$\omega_{01}$} & \textbf{$\phi_{01}$} & \textbf{$\kappa_{01}$} & \textbf{$v/H\delta$} \\
           & {(km)} & { (km)} & {(km)} & { (deg)} & { (deg)} & { (deg)} &  \\
    \midrule
    DS1024- & \multirow{2}[1]{*}{-0.187} & \multirow{2}[1]{*}{-3.909} & \multirow{2}[1]{*}{-0.652} & \multirow{2}[1]{*}{-0.020} & \multirow{2}[1]{*}{-0.202} & \multirow{2}[1]{*}{0.050} & \multirow{2}[1]{*}{0.016} \\
    1038DF096 &        &        &        &        &        &        &  \\
      DS1024- & \multirow{2}[1]{*}{0.037} & \multirow{2}[1]{*}{-3.167} & \multirow{2}[1]{*}{-2.307} & \multirow{2}[1]{*}{0.028} & \multirow{2}[1]{*}{-0.06} & \multirow{2}[1]{*}{0.019} & \multirow{2}[1]{*}{0.012} \\
    1038DF095 &        &        &        &        &        &        &  \\
    DS1049- & \multirow{2}[0]{*}{0.463} & \multirow{2}[0]{*}{-3.181} & \multirow{2}[0]{*}{0.453} & \multirow{2}[0]{*}{-0.15} & \multirow{2}[0]{*}{-0.002} & \multirow{2}[0]{*}{0.014} & \multirow{2}[0]{*}{0.017} \\
    2119DF055 &        &        &        &        &        &        &  \\
    \bottomrule
    \end{tabular}%
  \label{tab:IMCKH4}%
\end{table*}%

\subsection{Epipolar Resampling and the Accuracy of the DEM {(\textit{Mongolia Data})}}
\begin{figure}[]
\centering
  \includegraphics[width=\linewidth]{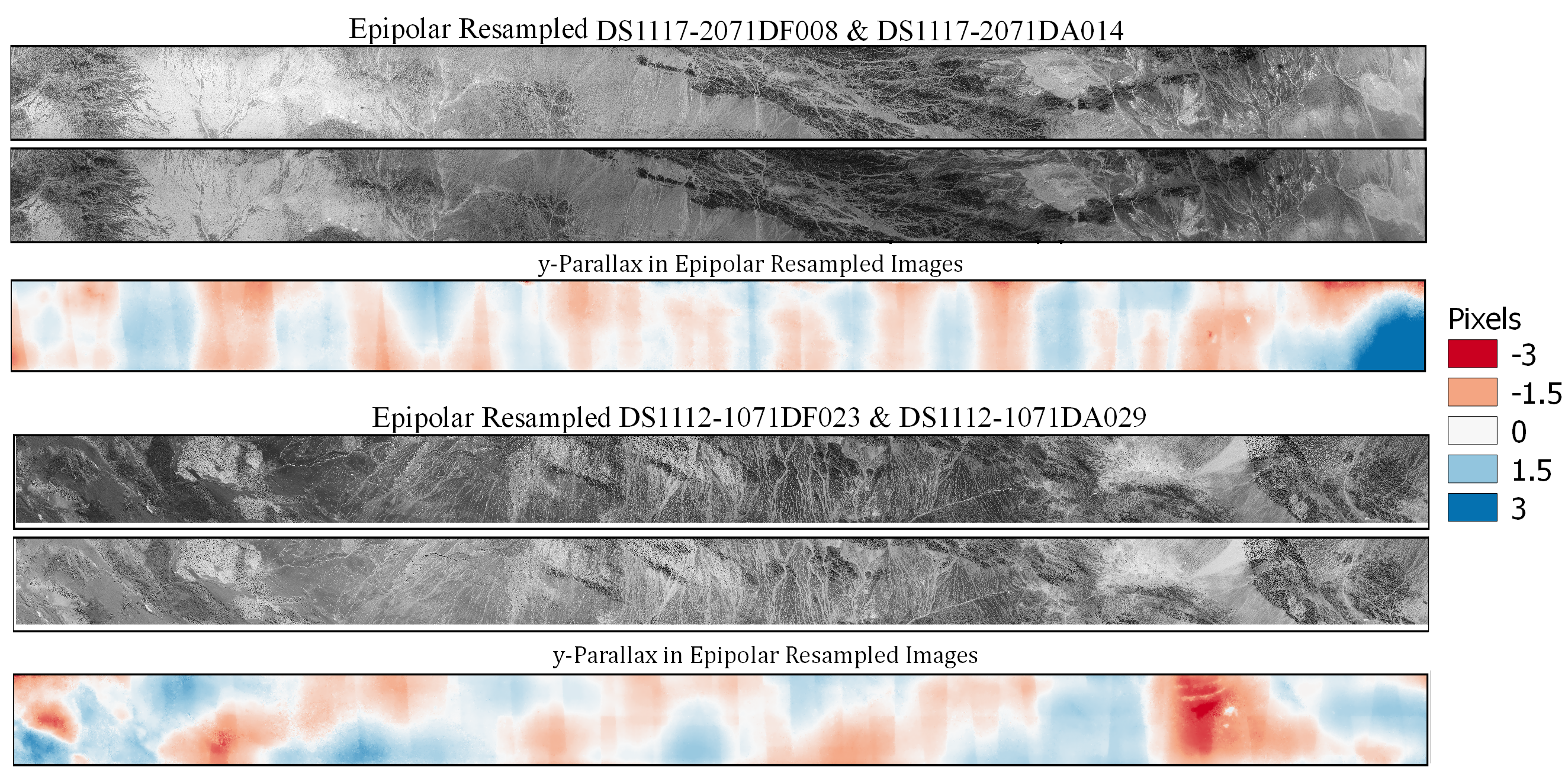}
  \caption{The two image pairs and the corresponding y-parallax in the epipolar resampled image pairs}
  \label{fig:EpipolarLinesRes}
\end{figure}

 The y-parallax in the epipolar resmapled images is computed using 2D correlation algorithm implemented in MicMac \cite{rosu2015measurement}. The SD of the y-parallax for the two rectified stereo pairs is 0.89 and 0.99 pixels. The y-parallax over majority of the area is less than $\pm 1$ pixels (Figure \ref{fig:EpipolarLinesRes}). However, certain regions with higher y-parallax of around 5 pixels are also observed. This can be due to the distortions in the film that affect locally certain part of the images. Such systematic errors correlate with the image residuals as shown in Figure \ref{fig:DistF023}.

\begin{figure}[]
\centering
  \includegraphics[width=\linewidth]{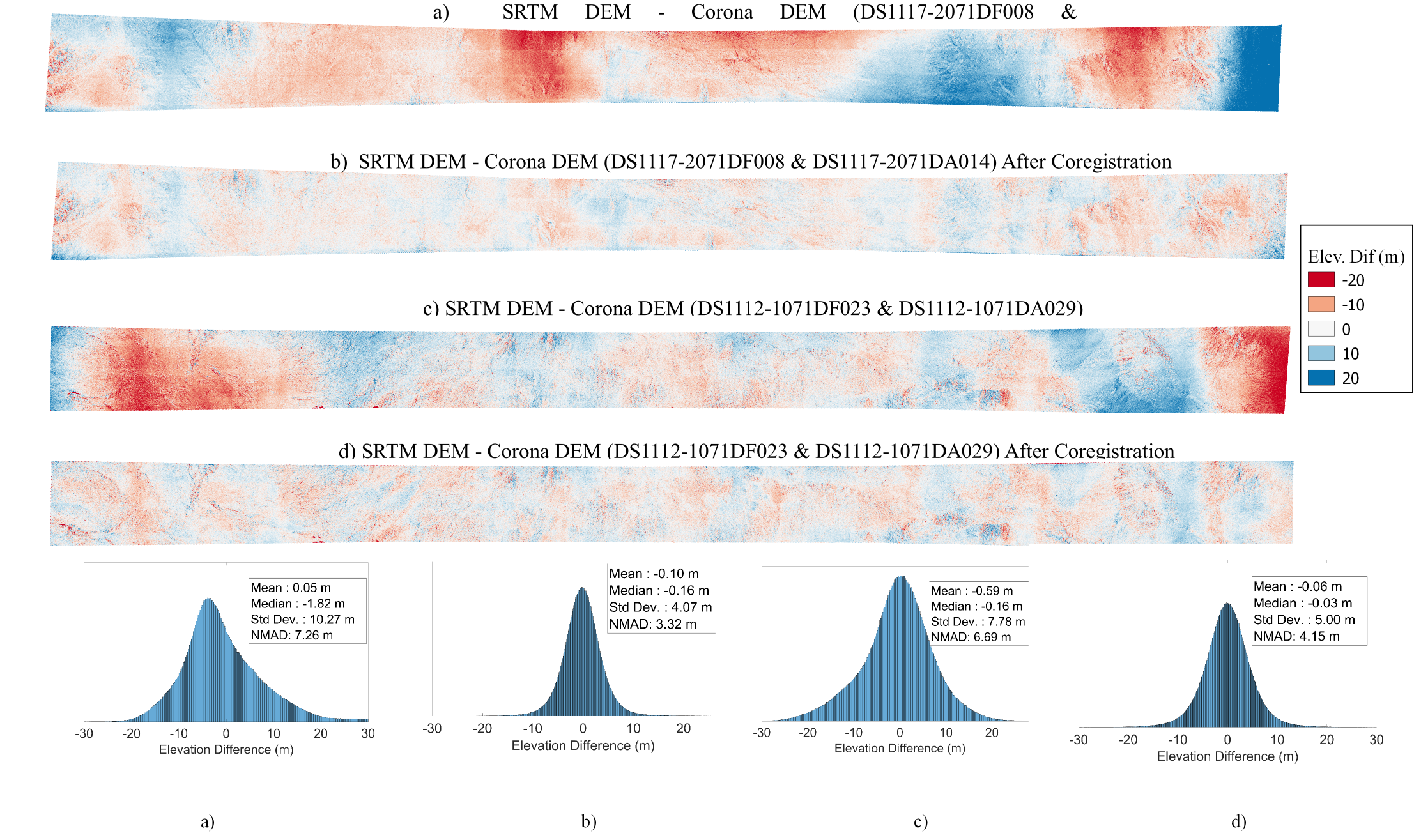}
  \caption{Elevation Differences between the Corona DEM and SRTM DEM for the two stereo pairs. The DEMs are divided in to patches and coregistered, which reduces the systematic errors in the Corona DEM.}
  \label{fig:DEMDif}
\end{figure}

 The NMAD of elevation differences between the two Corona DEMs and the SRTM DEM (\SI{30}{\m}) are \SI{7.26}{\m} and \SI{6.69}{\m} (Figure \ref{fig:DEMDif}). The elevation differences show systematic deviations of up to \SI{25}{\m} in the respective DEMs. To eliminate these systematic deviations in the Corona DEM, we further align tiles of the Corona DEMs with the reference DEMs by estimating a 3D affine transformation, {as explained in Section~\ref{subsec:dem_gen}}. This alignment of the DEMs reduces the NMAD of the elevation differences to \SI{3.32}{\m} and \SI{4.15}{\m} for the two Corona DEMs (Figure \ref{fig:DEMDif}). The size of each tile is approx. \SI{400}{\km \squared}, which is large enough to avoid fitting the 3D affine transform to the temporal surfaces changes that we are interested in estimating.

\subsection{Case Study: Glacier Change Estimation (\textit{Everest Data)}}
\begin{figure}[]
  \includegraphics[width=\linewidth]{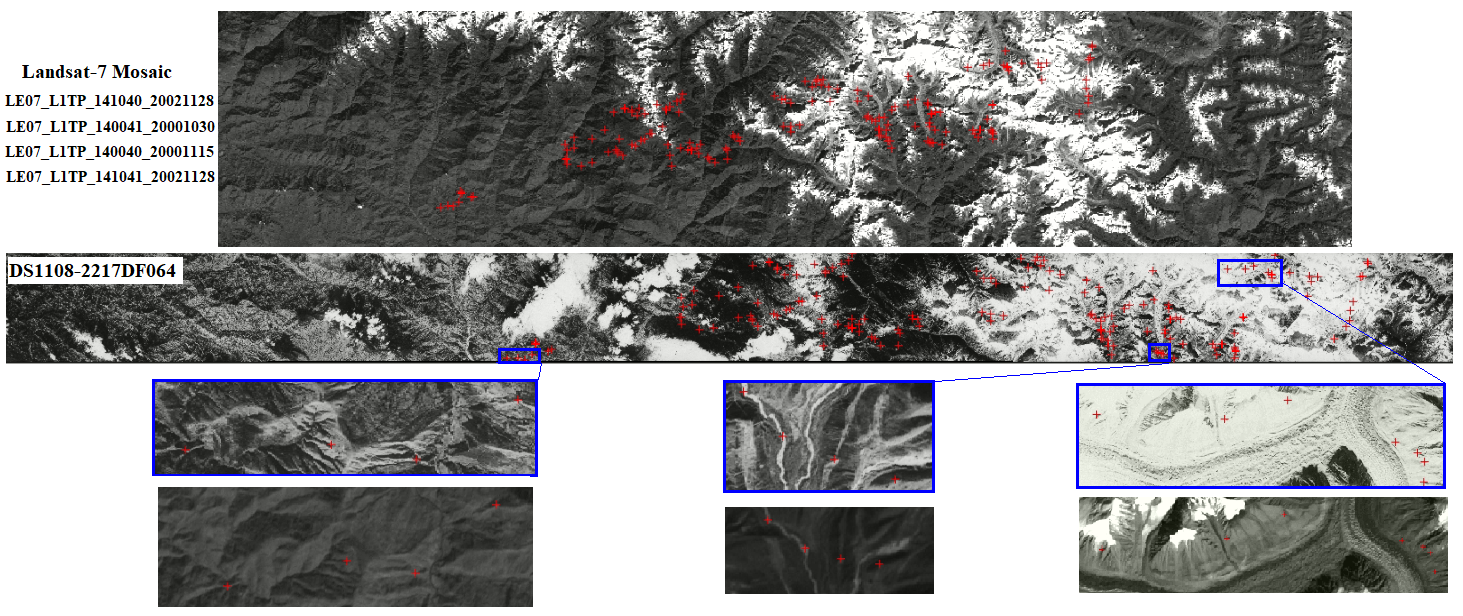}
  \caption{Feature points matched between 1969 Corona KH-4 and Landsat-7 panchromatic image  mosaic (year 2000) of the Everest region. Due to large area covered by the images, three locations are zoomed for visual comparison.}
  \label{fig:Everest_SuperGlue}
\end{figure}

In contrast to the scenes used for evaluating the camera model, the images over glacierized regions show significant seasonal and long-term variations. Due to difference in the Sun position and the cast shadows; the automatic feature matching across the mountainous terrain can be quite challenging. Even with these challenges, SuperGlue matched some points between Corona imagery and Landsat-7 ETM+ mosaic (Figure \ref{fig:Everest_SuperGlue}). However, no matches were found in the southwest part of the scene consisting of hilly forested terrain due to low texture and changes in the terrain and shadows.  

\begin{figure}[]
  \includegraphics[width=\linewidth]{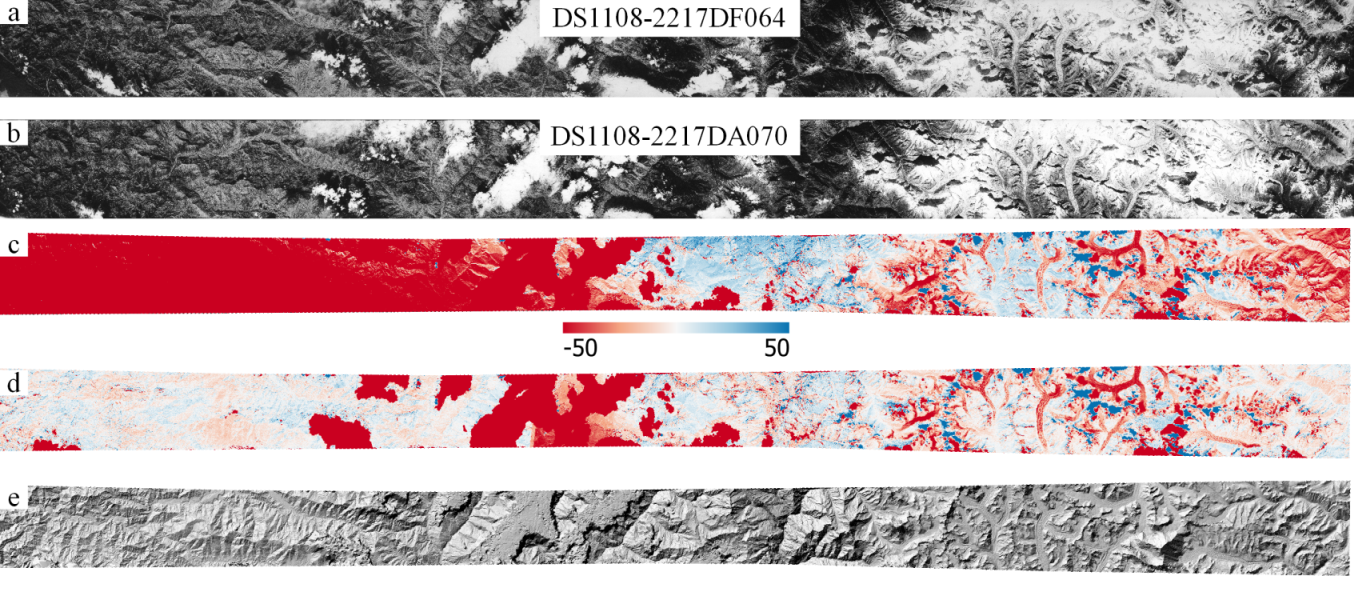}
  \caption{Corona stereo pair and the elevation differences between the Corona DEM and ALOS DEM (\SI{30}{\m}).  a) and b) show the Corona image pair c) shows the elevation difference with the ALOS DEM d) shows the elevation differences after tile based coregistration of the Corona DEM and the ALOS DEM e) shows the shaded Corona DEM.}
  \label{fig:Everest_Results}
\end{figure}

As no GCPs were available in one section of the image, the resulting DEM over that area show large elevation differences in comparison with the ALOS DEM \SI{30}{\m} (Figure \ref{fig:Everest_Results}). The Corona DEM is then coregistered with the ALOS DEM using smaller tiles, which removed the majority of the systematic elevation differences between the Corona and the ALOS DEM. The DEM coregistration is performed using the stable terrain i.e. the area without the glaciers. To filter out the glacierized part of the scene, we used the glacier polygons from the work of \cite{king2020six}. The Randolph Glacier Inventory (RGI) \cite{pfeffer2014randolph} polygons were manually modified to match glacier extent from 1969. The elevation differences show errors caused by textureless snow/ice cover in the accumulation zone of the glacier and steep rockwalls. Further large deviations are due to the presence of clouds in the Corona imagery (Figure \ref{fig:Everest_Results}).

\begin{figure}[]
  \includegraphics[width=1\linewidth]{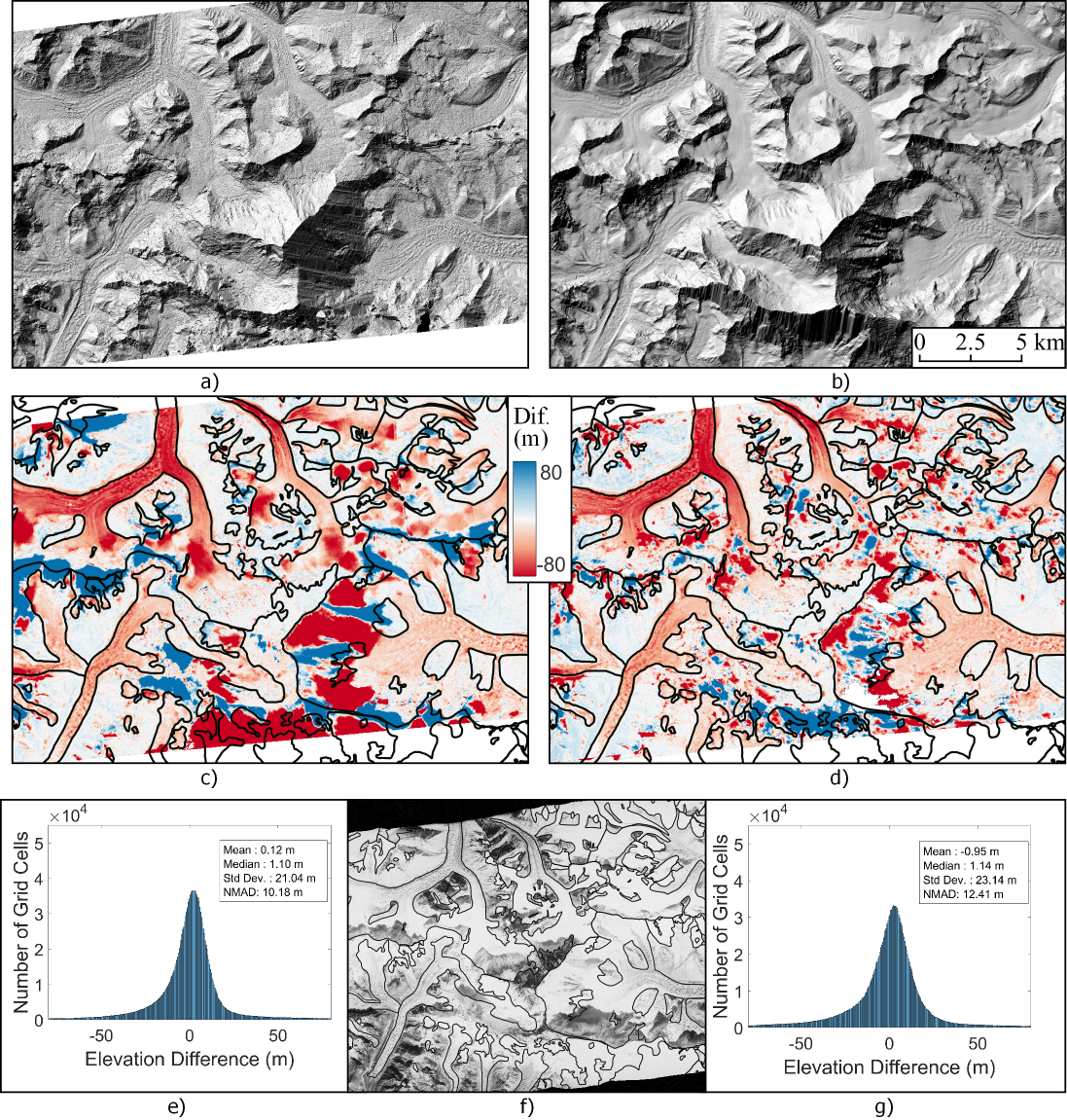}
  \caption{a) Hillshade of \SI{10}{\m} Corona DEM  b) Hillshade of \SI{10}{\m} Cartosat DEM (Date: 15/12/2018) c) Elevation differences between Cartosat  and Corona DEM computed using the proposed methodology d) Elevation differences between Cartosat  and Corona DEM computed using RSG \cite{king2020six} e) Histogram of elevation differences given in [c]  f) Corona orthophoto of the Everest Region g) Histogram of elevation differences given in [d] }
  \label{fig:Everest_Carto}
\end{figure}

The highest spatial resolution in the KH-4 missions is around \SI{1.8}{\m} at the center of the image and around \SI{3}{\m}  towards the edges of the film. Hence, DEMs with relatively high spatial resolution can be generated from the Corona images. Here, we compute a \SI{10}{\m} Corona DEM of the Everest region and compare it with a \SI{10}{\m} Cartosat DEM  of 15/12/2018 (Figure \ref{fig:Everest_Carto}) from the work of \cite{king2020six}. The elevation differences between the Corona and Cartosat DEMs show a decrease in the elevation of the glaciers and thus a negative glacier mass balance. The elevation differences over the stable terrain shows that the two DEMs are well coregistered.  A visual inspection of the shaded DEMs show similar level of details in the Corona and Cartosat DEMs over the well textured part of the image. Though the Corona DEM has higher noise level. The accumulation region of the glaciers' with no texture result in large elevation differences (Figure \ref{fig:Everest_Carto}). Note that  we generated spatially complete Corona DEM but in practice such unreliable elevation values can be filtered out to reduce the chances of introducing bias in the estimation of glacier volume change.

This \SI{10}{\m} Corona DEM of Everest region, generated using CoSP is compared with the Corona DEM generated using RSG as used in \cite{king2020six}. The NMAD of the elevation differences between CoSP Corona DEM and Cartosat DEM over the stable terrain is \SI{10.18}{\m}, which is slightly better than the NMAD of \SI{12.41}{\m} obtained by Corona DEM generated using RSG (Figure \ref{fig:Everest_Carto}). The elevation differences over the textured part of the glaciers are very similar for CoSP and RSG Corona DEMs.  However, the pattern of outliers in textureless accumulation regions of the glaciers show a very different pattern, which is possibly due to differences in regularization parameters of the dense matching algorithms used in CoSP and RSG. After performing the outlier removal and gap filling, the elevation difference from 1969 to 2018 over the glacierized area shown in Figure \ref{fig:Everest_Carto} is $-13.55\pm \SI{3.91}{\m}$ for CoSP and $-10.26\pm \SI{4.13}{\m}$ for RSG computed Corona DEMs. The elevation change for the Khumbu Glacier is $-16.96\pm \SI{3.91}{\m}$ for CoSP and $-17.20\pm \SI{4.13}{\m}$ for RSG generated Corona DEMs.

\subsection{Large scale Mapping using CoSP {(\textit{Pamir Data})}}
\begin{figure}[]
  \includegraphics[width=1\linewidth]{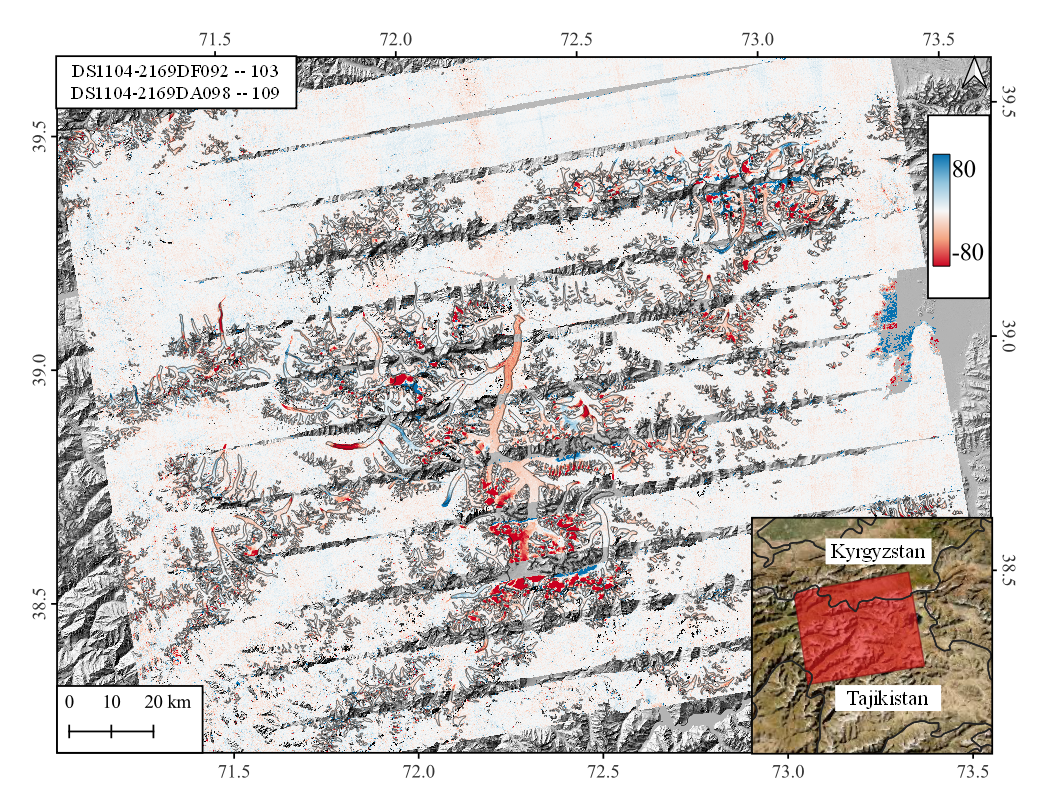}
  \caption{DEM differences between the Corona DEM computed using 12 Corona stereo pairs and the ALOS DEM over Pamir region.}
  \label{fig:Pamir}
\end{figure}

In order to show the applicability for large scale mapping, we apply the proposed methodology on a sequence of 12 Corona stereo pairs over Central Pamir (Figure \ref{fig:Pamir}). We applied all processing steps and differenced the resultant DEM with the \SI{30}{\m} ALOS DEM. The DEM differences show that CoSP works on a larger scale. The systematic elevation errors are low in magnitude over the entire area. The large number of surge type glaciers with either strong elevation loss in the middle reaches of the glacier and an elevation gain at its tongues or vice versa \cite{goerlich2020more}, can be well identified in the DEM differences. 

This sequence of Corona imagery also shows a potential complication in large scale processing. The consecutive images of the sequence have an overlap of approx. 6\% at the center of the format, which is less than the specified overlap of approx. 8\%. While the the fore and aft looking cameras have an overlap of approx. 90\%. As a result there are gaps in between the consecutive stereo pairs with greater voids towards the center of the image. These voids can be filled by considering the small overlap with the adjacent \textit{aft} image of the sequence. However, the coregistration of the DEMs with small width (e.g. $3 \times 200$ km) can be challenging in presence of unstable terrain. 

\section{Discussion}
\subsection{ Camera Model}
The results show that a SD ($\sigma_0$) of better than $2$ pixels in bundle adjustment can be obtained  with a rigorous Corona camera model consisting of time dependent collinearity equations. The image residuals show a pattern, which can be attributed to film distortions due to long term storage and large temperature variations during the mission as also observed for Hexagon KH-9 films \cite{surazakov2010positional, maurer2015tapping}. Furthermore, film bowing/buckling, misalignment in the components of the camera and atmospheric effects may also impart distortions in the image. It is pertinent to mention that these instruments were continuously upgraded and technical problems have occurred during the Corona missions, so there can be other factors associated with the residuals distortions observed in the Corona images. 

The rigorous Corona camera model used here is based on the work of \cite{sohn2004mathematical}. This camera model is more flexible as it includes six additional parameters for time dependent modelling of exterior orientation parameters. The estimated camera parameters are consistent with the expected values and the variation in their values is due to the systematic errors caused by film distortions. The estimated camera parameters show that the difference in the IMC mechanism in KH-4A and KH-4B is observed in the estimated camera parameters and the inclusion of IMC term in the camera model may not be necessary for KH-4B cameras.  

The height RMSE of check points obtained here i.e. \SI{5.79}{\m}  and \SI{8.42}{\m}  for the two Corona stereo pair (Table \ref{tab:resultsMong}) is higher than the height RMSE of check points of around \SI{4}{\m}  reported in \cite{sohn2004mathematical}. This decrease in the accuracy is expected due to the presence of film distortions, when considering the entire Corona stereo coverage (area $>$\SI{3000}{ \km\squared}) in comparison to using certain section of the image (area of \SI{561}{\km \squared}) as done in \cite{sohn2004mathematical}. As Figures  \ref{fig:DistF023} and \ref{fig:EpipolarLinesRes} suggest, residuals distortions may have a higher magnitude in certain parts of the image and considering larger image area may decrease the accuracy of check points due to these distortions. The patch based fine coregistration with the reference DEM improves the accuracy and reduces the SD ($\sigma$) of elevation differences to around \SIrange{4}{5}{\m} (Figure \ref{fig:DEMDif}). This shows that accuracy similar to \cite{sohn2004mathematical} can be obtained over the entire stereo coverage as fine coregistration compensates the effect of the distortions along the length of the film to a certain extent. 

\cite{shin2008rigorous} have reported a height RMSE of \SI{12.34}{\m} for checkpoints, which is higher than RMSE obtained in our work. However, they used a scan with \SI{12}{\um} pixel size, which reduces accuracy in comparison to \SI{7}{\um} pixel size used in this work. The camera model in \cite{shin2008rigorous} was based on time dependent collinearity equations but they only modeled motion along flight direction as an additional time dependent parameter. 

These results further show an improvement in accuracy in comparison to the work of  \cite{bhattacharya2021high}, who have reported an RMSE of around $2.5$ pixels for RSG Corona camera model. In addition, the statistics of the elevations differences with a \SI{10}{\m} Cartosat DEM also show slight improvement in comparison to the Corona DEM generated using RSG in the work of \cite{king2020six} (see Figure \ref{fig:Everest_Carto}). This demonstrates that CoSP can achieve  accuracy similar to the best reported accuracy in the previous work on Corona imagery, while considering the entire Corona stereo image coverage. 


\subsection{Corona Data Limitations}
\subsubsection{Film Reference Data}
The reference metadata exposed on the film is important for reconstruction of the panoramic geometry and correction of film distortions. However, as shown earlier, the PG rail holes show unexpected deviations and are not feasible for correction of film distortions or reconstruction of film geometry. The KH-4 and KH-4A imagery doesn't have PG stripes and PG rail holes. Furthermore, we have observed several KH-4B images in which PG stripes are not fully visible.  So, bending correction as performed here may not be applicable to all Corona images, which will have an effect on the expected accuracy. 

\subsubsection{Low Texture and Film Saturation}
One limitation of Corona imagery, which is especially significant for glacier mass balance studies is the low texture and contrast over snow covered part of the glaciers mainly caused by the saturation due to light exposed on the film. This issue of textureless area has a greater impact in snow covered parts of the glaciers especially in the accumulation regions, where it may be difficult to estimate reliable elevation data \cite{goerlich2017glacier}. This problem is further exacerbated by lack of cloud free images in the late summer to early winter time period. Although the problem of low image texture over snow covered area is common to all optical sensors, the spatial resolution, wavelength range, radiometric resolution, dynamic range and image acquisition time can significantly impact the quality of DEM over the snow covered areas.   

\subsubsection{Inconsistent Overlap}
There are two overlaps to consider, when creating DEMs from a sequence of Corona stereo images: First is the overlap between the successive frames of a Corona camera and the second overlap is between the stereo pair. The overlap between the successive frames is specified to be approx. 8\% at the center of the format and is directly related to the camera scan rate, which in turn depends on satellite velocity to height ratio \cite{NROJ3}. The second overlap is between the stereo pair i.e. images of an area from the \textit{fore} and \textit{aft} looking cameras. This overlap is typically around 90\% but varies over different acquisitions. The overlap between the successive images is rather small, which results in a small overlap between the DEMs from the successive stereo pairs \cite{goerlich2017glacier, jacobsen2020calibration}. Hence, deviations from the specified scanning rate or attitude perturbations can cause gaps in the DEMs as seen in the image sequence over Pamir, where the overlap between the successive images was approx. 6\%.


\subsubsection{Scanning Artifacts}
\begin{figure}[]
  \includegraphics[width=\linewidth]{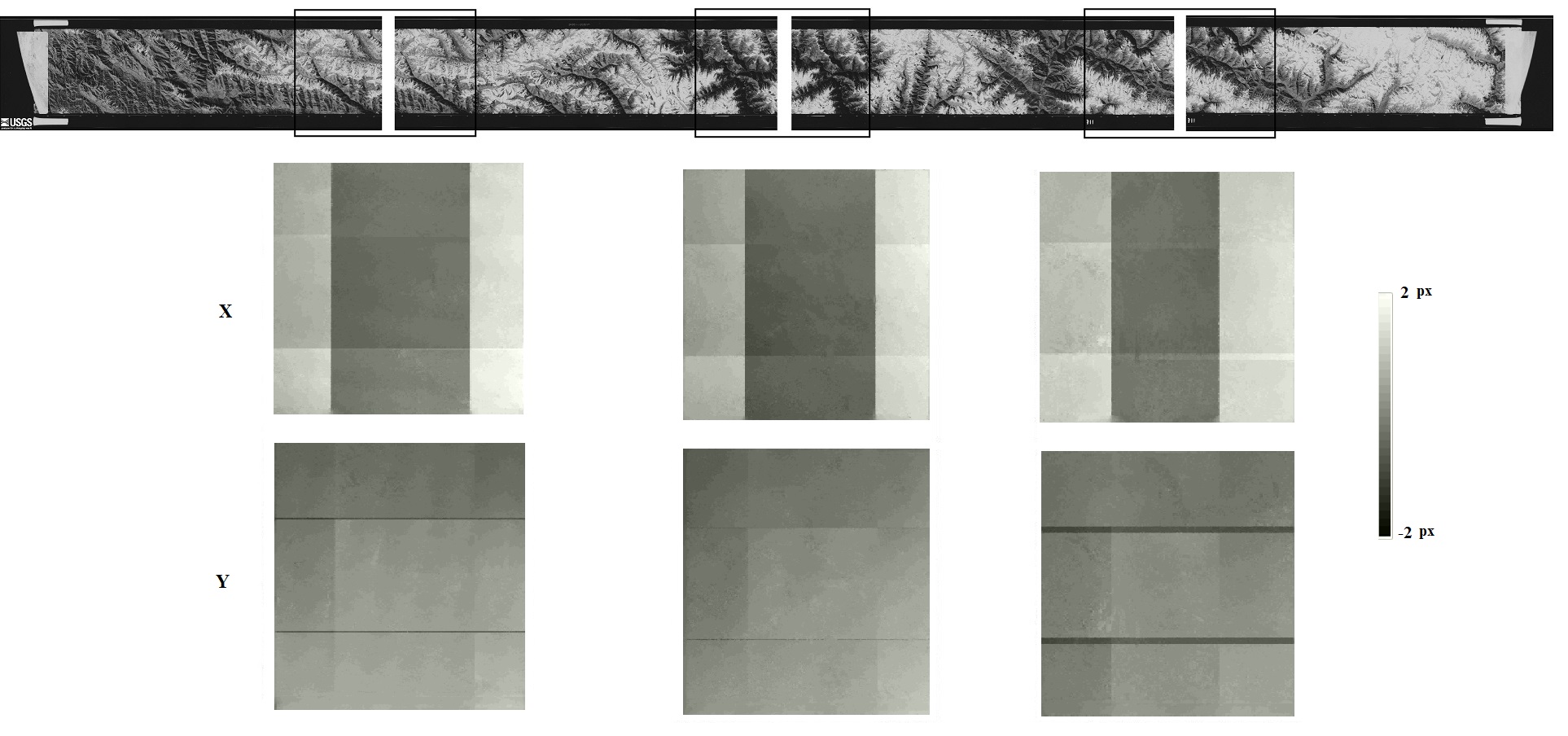}
  \caption{Scanning artifacts observed in the aligned overlapping parts of the scans. The shift along the x and y axis are estimated using 2D cross correlation in MicMac.}
  \label{fig:ScanArt}
\end{figure}

The scanned imagery delivered by USGS has been reported to contain scanning artifacts \cite{lauer2019exploiting,galiatsatos2007high,gheyle2011scan,dehecq2020automated}. These artifacts occur due to error in the scanner calibration. After processing KH-4 imagery of different missions, we have also observed such artifacts in the scanned Corona images.  These blocking artifacts can be observed as a misalignment in the overlapping stitched parts of the scans. To quantify the misalignments, we first align the individual scans by a rotation and translation using the tie points extracted in the overlapping region of the neighboring scans. Then dense image matching is performed on the aligned images. The result of the matching of the three image parts along $x$ and $y$ directions are shown in Figure \ref{fig:ScanArt}. As the Corona images have been available since long, these images may have been scanned with a different scanner or with varying scanner calibration. As a consequence, these artifacts show varying magnitude in different images. Such blocking artifacts are also visible in the DEM difference shown in Figures \ref{fig:EpipolarLinesRes} and \ref{fig:DEMDif}.

\subsection{GCP Generation}
The main advantage of CoSP is the automated processing of Corona imagery, which enables large scale mapping using Corona imagery. Automatic GCP generation is an essential part of this pipeline. Our main focus of application has been on the glaciers of high mountain Asia and GCP generation using SuperGlue has largely been successful in generating automatic GCPs over such terrain. However, automatic GCP generation may still be challenging over scenes dominated by  large forested areas, agricultural land, large water bodies, deserts and urban areas, which may have changed significantly over a period of time. Performing multi image bundle adjustment for a sequence of Corona images can potentially resolve the issue of unavailability of GCPs in some  images of the sequence if GCPs can be extracted for other images of the sequence. Another possible solution is to utilize DEMs to matching features for georeferencing of the historical imagery as done in \cite{zhang2021feature,maurer2015tapping}.

\subsection{DEM Coregistration}
Here, we used a tile based approach to coregister the Corona DEM with a reference DEM, which appears to work well in the above test cases. However, problems in coregistration can appear if large parts of the scene contain unstable terrain, outliers or data voids. Thus, a coregisteration using larger area or whole DEM will be more suitable. A linear transformation in 3D was only sufficient to coregister  small parts (approx. 10\%) of the scene and a nonlinear transformation will be required for coregistration of larger areas.

There are other approaches to DEM cogregistration, which apply corrections functions to a raster DEM. Such approaches have widely been used for coregistration of DEMs for quantification of glacier elevation change \cite{nuth2011coreg, pieczonka2015region, girod2017mmaster, hugonnet2021accelerated}. These approaches typically consist of multiple components such as correction of DEM shift, elevation dependent bias correction, sensor specific corrections such as the effect of jitter in ASTER DEMs and correction of remaining systematic elevation differences by polynomial fittting.  Such approaches may also be investigated for improvement of the DEM coregistration strategy in CoSP.

\section{Conclusions}
\label{ch:Conclusions}
This work presented CoSP: A pipeline for automated processing of Corona KH-4 series imagery. A rigorous camera model with modified collinearity equations consisting of time dependent exterior orientation parameters was used to model the panoramic image acquisition geometry of the Corona cameras.  A deep learning-based feature extractor was used to automatically match feature points between Corona imagery and medium  to high resolution satellite imagery. These feature points were used as GCPs to evaluate the accuracy of the camera model. The results show that overall a $\sigma_0$ better than 2 pixels can be achieved, while considering the entire Corona image. The estimated camera parameters are consistent with the known parameters as well as the IMC mechanism on the Corona camera system. Systematic image residuals up to six pixels have been observed in different Corona scenes. These errors are perhaps due to film distortions that have occurred during the mission and the storage in the 50-year period. Consequently, the resulting Corona DEMs exhibit systematic elevation differences of up to \SI{25}{\m}, which could be reduced by a tile-based coregisteration of the Corona DEM. A tie point based epipolar resampling of the Corona images showed residual y-parallax of $\pm 1$ across majority of the rectified image pair. The accuracy of the derived 3D data was further improved by performing bending correction using the PG stripes exposed on the film. Finally, it was shown that the proposed methodology can be used on Corona scenes over high relief and glacierized terrain, which enables calculation of multi-decadal glacier mass balance over large areas.

\bibliographystyle{unsrtnat}
\bibliography{references}  






\end{document}